\PassOptionsToPackage{prologue,dvipsnames,table}{xcolor}
\documentclass[sigconf]{acmart}
\AtBeginDocument{%
  \providecommand\BibTeX{{%
    \normalfont B\kern-0.5em{\scshape i\kern-0.25em b}\kern-0.8em\TeX}}}

\usepackage{multirow}
\usepackage{algorithm}
\usepackage{algorithmic}
\usepackage{tcolorbox}
\usepackage{subfigure}

\usepackage{amsmath,amsfonts,bm}

\def\eqref#1{equation~\ref{#1}}
\def\1{\bm{1}}

\DeclareMathAlphabet{\mathsfit}{\encodingdefault}{\sfdefault}{m}{sl}
\SetMathAlphabet{\mathsfit}{bold}{\encodingdefault}{\sfdefault}{bx}{n}

\def\gE{{\mathcal{E}}}

\def\gG{{\mathcal{G}}}

\def\gL{{\mathcal{L}}}

\def\gO{{\mathcal{O}}}

\def\gR{{\mathcal{R}}}

\def\gT{{\mathcal{T}}}

\def\gV{{\mathcal{V}}}

\def\gZ{{\mathcal{Z}}}

\newcommand{\KL}{D_{\mathrm{KL}}}

\DeclareMathOperator*{\argmax}{arg\,max}

\newtheorem{definition}{Definition}

\begin{document}

\settopmatter{printacmref=false}
% \renewcommand\footnotetextcopyrightpermission[1]{} % removes footnote with conference 
% \pagestyle{plain}

% \title{Planned Retrieval of Semi-structured Data for Question Answering with Textual and Relational Information}
\title{PASemiQA: Plan-Assisted Agent for Question Answering on Semi-Structured Data with Text and Relational Information}

\author{Hansi Yang}
\email{hyangbw@cse.ust.hk}
\affiliation{%
  \institution{CSE, HKUST}
  % \streetaddress{P.O. Box 1212}
  \city{Hong Kong}
  \country{China}
}

% \affiliation{%

\author{Qi Zhang}
\affiliation{%
  \institution{Ant Group}
  \city{Beijing}
  \country{China}
}

\author{Wei Jiang}
\affiliation{%
  \institution{Ant Group}
  \city{Beijing}
  \country{China}
}

\author{Jianguo Li}
\affiliation{%
  \institution{Ant Group}
  \city{Beijing}
  \country{China}
}

% \affiliation{%

% \affiliation{%

% \affiliation{%

% \affiliation{%

% \affiliation{%


\begin{abstract}
% Despite its ability in answering questions from different domains, 
% existing large language models (LLMs) may suffer from the hallucination issue when answering questions that require professional and up-to-date knowledge. 
% Such limitation calls for the development of retrieval-augmented generation (RAG) techniques, 
% which retrieves relevant information from external data sources 
% and utilizes retrieved contents to answer the given question. 
% Nevertheless, existing RAG methods often focus on a single type of database 
% (e.g., vectorized text database or knowledge graphs) 
% and cannot handle real-world questions on semi-structured data that contains both text and relational information.  
% % Nevertheless, existing RAG methods cannot jointly leverage information storing in different formats, e.g., relational databases and text data sets, 
% % which is naturally present in diverse domains. 
% Motivated by such limitations, 
% we propose a novel method PASemiQA to jointly utilize text and relational information in semi-structured data to answer questions. 
% PASemiQA first obtains a plan to determine text and relational information in semi-structured data
% that is useful to answer the given question. 
% Then based on the generated plan, 
% we use an LLM agent to traverse the semi-structured data and
% extract related information for this question. 
% % extracts related entities from the given query 
% % based on both keywords and text embeddings. 
% Empirical results across semi-structured data sets from different domains demonstrate the effectiveness of the proposed method. 
Large language models (LLMs) have shown impressive abilities in answering questions across various domains, but they often encounter hallucination issues on questions that require professional and up-to-date knowledge. To address this limitation, retrieval-augmented generation (RAG) techniques have been proposed, which retrieve relevant information from external sources to inform their responses. However, existing RAG methods typically focus on a single type of external data, such as vectorized text database or knowledge graphs, and cannot well handle real-world questions on semi-structured data containing both text and relational information. To bridge this gap, we introduce PASemiQA, a novel approach that jointly leverages text and relational information in semi-structured data to answer questions. PASemiQA first generates a plan to identify relevant text and relational information to answer the question in semi-structured data, and then uses an LLM agent to traverse the semi-structured data and extract necessary information. Our empirical results demonstrate the effectiveness of PASemiQA across different semi-structured datasets from various domains, showcasing its potential to improve the accuracy and reliability of question answering systems on semi-structured data.
\end{abstract}

\begin{CCSXML}
<ccs2012>
   <concept>
       <concept_id>10002951.10003317.10003338</concept_id>
       <concept_desc>Information systems~Retrieval models and ranking</concept_desc>
       <concept_significance>300</concept_significance>
       </concept>
   <concept>
       <concept_id>10002951.10003317.10003347.10003348</concept_id>
       <concept_desc>Information systems~Question answering</concept_desc>
       <concept_significance>500</concept_significance>
       </concept>
   <concept>
       <concept_id>10010147.10010178.10010187</concept_id>
       <concept_desc>Computing methodologies~Knowledge representation and reasoning</concept_desc>
       <concept_significance>100</concept_significance>
       </concept>
 </ccs2012>
\end{CCSXML}

\ccsdesc[300]{Information systems~Retrieval models and ranking}
\ccsdesc[500]{Information systems~Question answering}
\ccsdesc[100]{Computing methodologies~Knowledge representation and reasoning}

\keywords{Large Language Model, Question Answering, Graph-structured Data}



\maketitle

% {\bf Relevance to the Web: this work focuses on improving the Web as a technical infrastructure. We propose a novel method to better utilize semi-structured data available on the Web for answering questions using LLM.}
% Methods to enhance, augment, integrate or synergize semantic models such as knowledge graphs and LLMs to enlarge the functionality of either or both models
% interested in knowledge graphs and other forms of structured data models with machine-interpretable semantics that are being widely adopted for many advanced applications on the Web. In this track, we invite original research submissions related to methods, algorithms, techniques and applications supporting the creation, acquisition, publication and consumption of 
% interlinked structured data corpora – available on the web
% , the (human-assisted) semantic integration, the enrichment and processing of large, real-world datasets in a Web context.

\section{Introduction}

% Powerful LLMs have wide applications. 
Recent years have witnessed the emergence of powerful large language models (LLMs) 
obtained from large-scale pre-training on vast amount of corpus~\cite{brown2020language, achiam2023gpt, yang2024qwen2, dubey2024llama, touvron2023llama, team2023gemini, caruccio2024claude}. 
Despite their strong language understanding and instruction following ability~\cite{wei2022finetuned}, 
directly using LLMs for downstream applications may encounter the hallucination issue~\cite{ji2023survey}, 
where LLM randomly generates an answer that satisfies semantic rules 
but does not match the reality. 
Such issue makes it not accountable to directly use LLM on accuracy-critical applications, 
and motivates the development of retrieval-augmented generation (RAG) techniques~\cite{Zhao2024RetrievalAugmentedGF} 
that try to utilize accurate and up-to-date external information 
to alleviate the hallucination issue. 
A common RAG pipeline is to first retrieve relevant contents from a given database based on the input, 
and let the LLM model generate its output from the original input and retrieved content together. 
% first utilizes the language understanding ability of LLM to retrieve related contents based on the query from a given database, 
% Then based on the retrieved contents, 
% we can use LLM to generate answers with its language understanding abilities. 
RAG techniques have found application in diverse domains, 
including molecular generation for drug discovery~\cite{wang2023retrievalbased}, 
biomedical literature understanding~\cite{frisoni2022bioreader}, 
and code completion within a repository~\cite{zhang-etal-2023-repocoder}. 

\begin{figure*}[ht]
  \centering
  \subfigure[(Vectorized) text database. Example content is shown in the blue box. \label{fig:text}]{\fbox{\includegraphics[width=0.31\textwidth]{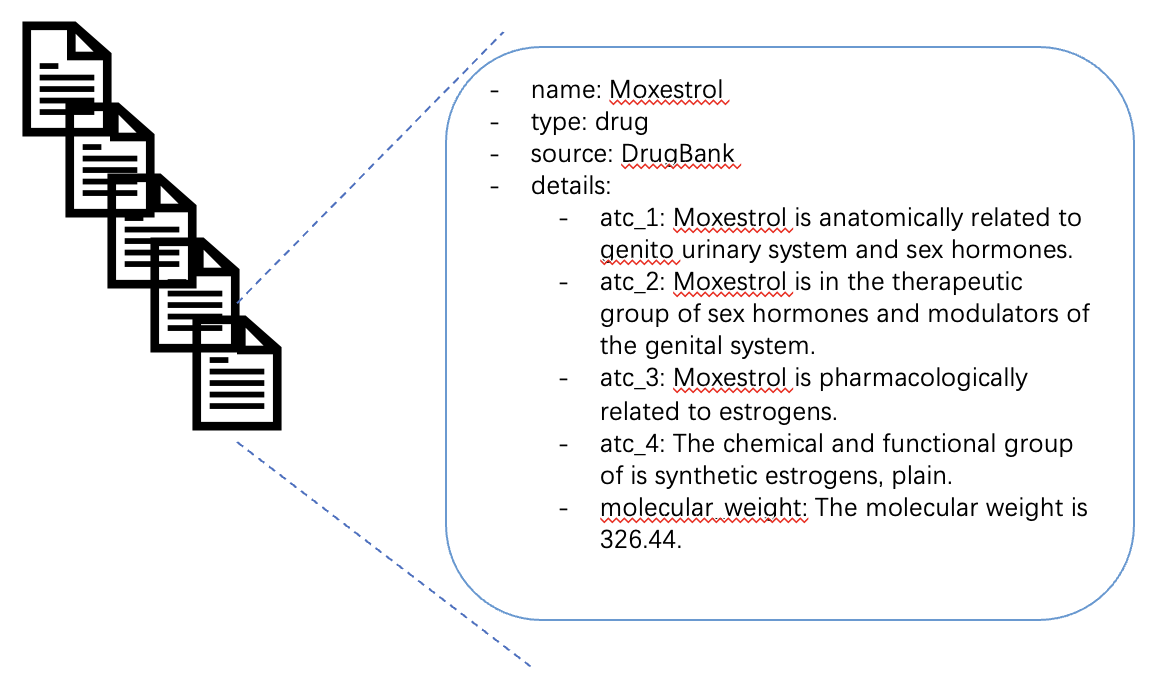}}}
  \subfigure[Knowledge graph (KG). Edge types (relations) are shown near each edge. \label{fig:kg}]{\fbox{\includegraphics[width=0.32\textwidth]{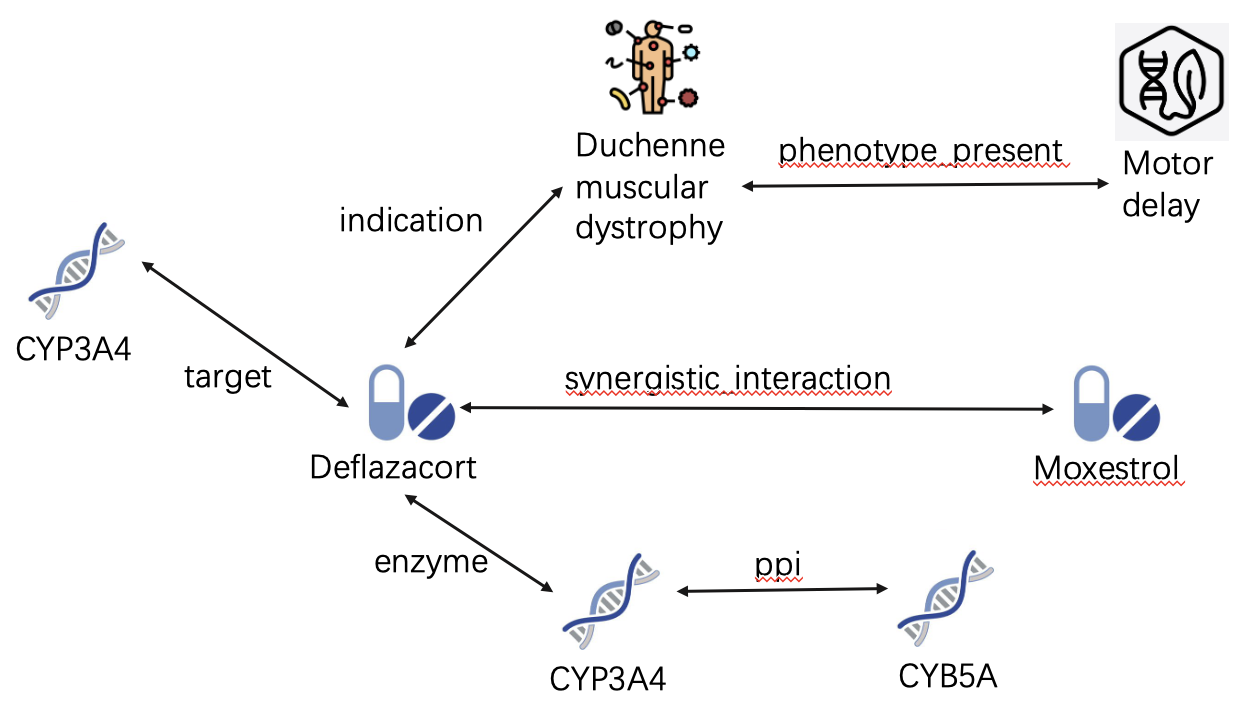}}}
  \subfigure[Semi-structured data. Node text descriptions are shown in corresponding blue boxes. Edge types (relations) are shown near each edge. \label{fig:semi}]{\fbox{\includegraphics[width=0.32\textwidth]{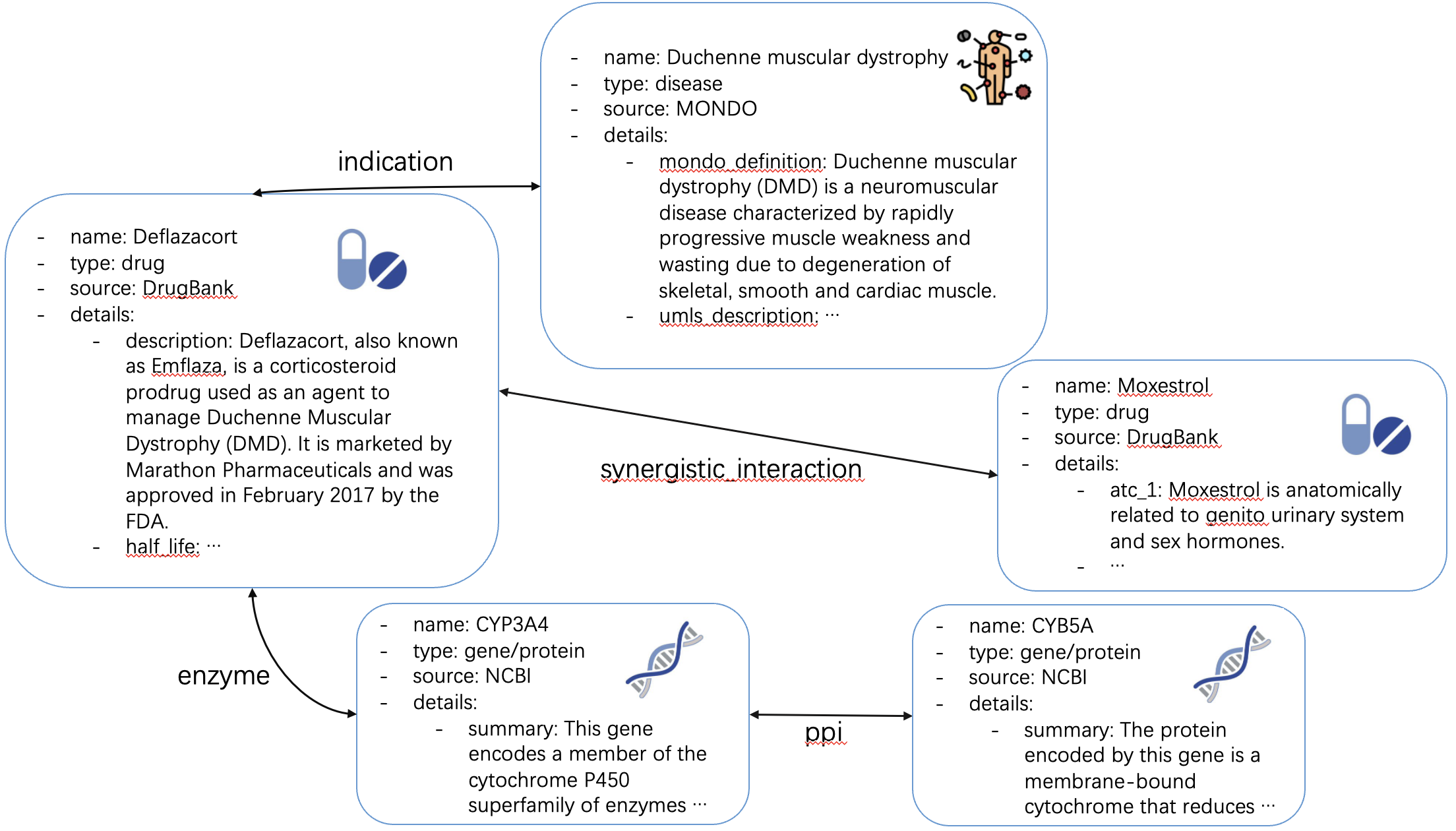}}}
  \vspace{-5px}
  \caption{Comparison of different data structures to supplement question answering of LLM.}
  \label{fig:data}
\end{figure*}

Based on the type of database that stores relevant contents for retrieval, 
existing RAG methods can be categorized into several types. 
The most common type is based on vectorized document database~\cite{yao2023react, shi-etal-2024-replug,zhang-etal-2023-repocoder}, 
where we retrieve text chunks based on their relevance to the input. 
% Since documents are stored separately, 
Such approach requires all information be stored in text format, 
and cannot perform exact step-by-step reasoning 
since the text chunks are separately retrieved based on the input. 
% based on text information in this database. 
Another common approach is based on knowledge graphs (KG~\cite{ji2022kg}), 
also referred as KGQA methods~\cite{sun2024thinkongraph, luo2024reasoning, xu2024generate}. 
An example of KG is shown in Figure~\ref{fig:kg}, where nodes correspond to entities in the real world, 
and edges between nodes indicate relations between pairs of entities. 
% Here, we retrieve information from a graph 
% whose 
While KG allows easy implementation of step-by-step reasoning 
by traversing along a path connecting different entities in KG, 
its format may also restrict the expressive power, 
as an edge connecting two entities cannot express complex relations that can cover multiple entities. 

To overcome the above limitations of each data modality, 
a straight-forward idea is to combine these two data modalities together, 
which leads to so-called semi-structured data~\cite{wu2024stark}. 
It combines knowledge graph with node-level documents to allow more flexible information storing. 
Due to its flexibility, semi-structured data  naturally find existence in diverse domains. 
An example in biomedical domain is shown in Figure~\ref{fig:semi}, 
where we use nodes to represent different biomedical objects, 
edges describe interactions between them, 
and each node can have additional documents as the description for the corresponding biomedical object. 
Another example is from e-commerce platforms, 
where products correspond to nodes with rich text information such as description and reviews, 
and connections between products can indicate the co-purchasing behaviors recorded by the platform. 
Despite such wide application, 
limited works consider how to answer questions with semi-structured data. 
Existing baseline methods~\cite{wu2024stark} simply borrow the baseline methods for RAG, 
which can neglect complex connections between nodes in semi-structured data, 
and perform badly when the question present includes complex reasoning on different relations across nodes in the semi-structured data. 

% Motivated by the above limitations, in this paper, we propose a novel method PASemiQA (Plan-Assisted Question Answering with Semi-structured data) that jointly utilize text and relational information to answer questions.
% We first note that different to standard RAG or KGQA methods, 
% answering questions with semi-structured data needs 
% to identify both text and relational information included in the question, 
% which cannot be well handled by existing works. 
% As such, we propose a novel planning module to select entities and relations 
% that are useful to answer the given question. 
% % for a given question based on the question and text embedding, 
% % which covers entities and relations mentioned by this question 
% % that should be useful for accurate answering this question. 
% Based on the generated plan, 
% we design an agent framework 
% that can be implemented by an LLM to traverse the semi-structured data and 
% extract relevant information for this question. 
% % extracts related entities from the given query 
% % based on both keywords and text embeddings. 
% Empirical results across semi-structured data sets from different domains demonstrate the effectiveness of our proposed method. 

Motivated by the limitations of existing approaches, we propose PASemiQA (Plan-Assisted Question Answering with Semi-structured data), 
a novel method that leverages both text and relational information to answer questions. Unlike standard RAG or KGQA methods, our approach tackles the unique challenge of identifying both text and relational information from the question. To address this, we introduce a novel planning module that selects relevant nodes and edges to inform the question-answering process. This plan is then used to guide an agent framework, implemented by an LLM, to traverse the semi-structured data and extract relevant information. Our method's effectiveness is demonstrated through empirical results on different semi-structured datasets from various domains.

Our contribution can be listed as follows:
\begin{itemize}
\item We propose a novel planning module that can generate informative plans for answering questions with both relational and text information. 
\item We propose an agent framework that can traverse the semi-structured data based on the generated plan to extract related information for this question. 
    \item We demonstrate the effectiveness of our proposed method through empirical results on different semi-structured datasets from various domains.
\end{itemize}

\section{Related Works}

\subsection{Retrieval-Augmented Generation (RAG)}

Generally, retrieval-augmented generation (RAG) methods~\cite{Zhao2024RetrievalAugmentedGF} consist of two parts: 
retrieval, which finds relevant information from external data sources for the given input, 
and generation, which generates the output based on retrieved information and the original input. 
A simple yet effective way of retrieval is to use text embedding similarity, 
which selects text chunks whose text embeddings have the largest cosine similarity to the text embedding of the input. 
Some later works propose to improve the retrieval process to be more effective than simply comparing text embeddings. 
For example, ReAct~\cite{yao2023react} uses multiple round of retrieval enabled by Chain of Thought (CoT~\cite{wei2022cot, zhang2023automatic, chu-etal-2024-navigate}) techniques
to retrieve more relevant information for the input. 
REPLUG~\cite{shi-etal-2024-replug} proposes to treat the generation process as a black box, 
and fine-tune a retrieval model that can lead to better final results of generation. 

The most simple way to generate the output is to directly fed the retrieved contents into a LLM model 
and use it to generate the output based on the original input and retrieved content. 
RETRO~\cite{pmlr-v162-borgeaud22a} further proposes to fine-tune the LLM on the desired output based on the retrieved data 
to improve the quality of generated content.  
% Some works also integrate CoT techniques to enhance the generation process. 
% \cite{Trivedi2022InterleavingRW} uses CoT to decompose the RAG process into multiple steps to generate more reasonable output, 
% while 
Another work~\cite{li2024chainofknowledge} uses CoT to break down complex problems into multiple sub-problems based on the retrieved content 
before generating the answer. 
% These approaches have improved the effectiveness of RAG in handling complex tasks.

Despite the above works that focus on vectorized text databases, 
some recent works have considered retrieving contents from a graph database. 
For example, G-retriever~\cite{he2024g} utilizes RAG techniques on a text-attributed graph, 
where nodes and edges are described by a short piece of text, 
and selects relevant node and edge in this graph to answer the given question. 
% The approach of G-retriever is quite similar to KGQA methods introduced below, 
% and still does not tackle the problem of jointly utilizing text and relational information. 
Nevertheless, the considered text-attributed graph only has small amount of text information, 
and G-retriever cannot be applied to semi-structured data where each node can have large amount of text information. 
% which makes G-retriever quite similar to KGQA methods introduced below 
% but does not fit RAG pipeline. 
GraphRAG~\cite{edge2024local} proposes a RAG pipeline on a set of text chunks that are connected based on their semantic similarities. 
Based on such connections, GraphRAG can answer complex questions that may involve summarizing across different text chunks. We note that this approach is also distinct from our method as we need to utilize external relational information 
that is not present in GraphRAG. 

\subsection{Knowledge Graph Question Answering (KGQA)}

% Some previous works [] attempt to train KG embeddings to predict answers by similarity scores under incomplete KG. Compared to these previous KGE-based works, we propose leveraging LLMs for QA under incomplete KG to study whether LLMs can integrate internal and external knowledge well. 

Mathematically, a knowledge graph (KG~\cite{ji2022kg, wang2017kg}) can be defined as a set of triplets $(s,r,o)$, 
where $s$ and $o$ are two entities, and $r$ is the relation between these two entities. 
Various methods have been proposed to utilize the language understanding ability of LLM to utilize relational information in a knowledge graph to answer questions. 
Based on the underlying mechanism, 
these methods can be classified into two categories: 
Semantic Parsing (SP) methods~\cite{li-etal-2023-shot} and Retrieval Augmented (RA) methods~\cite{sun2024thinkongraph, luo2024reasoning}. 
SP methods first transform the question expressed by natural language to a structural query using LLMs. 
Such queries can then be directly executed on a knowledge graph to directly derive the answers. 
An example SP method is KB-BINDER~\cite{li-etal-2023-shot}, 
which first generates logical forms from the given question, 
and then binds the logical forms to executable structural queries with entity and relation binders. 
However, the effectiveness of these methods relies heavily on the quality of the generated queries and the completeness of KGs. 
RA methods share more similarities to RAG methods, which also retrieve related information from the KG to generate accurate answers for the given querstion~\cite{sun2024thinkongraph, xu2024generate}. 
ToG~\cite{sun2024thinkongraph} treats the LLM as an agent to interactively explore relation paths step-by-step on KGs and perform reasoning based on the retrieved paths. 
RoG~\cite{luo2024reasoning} first generates relation paths as faithful plans, and then use them to retrieve valid reasoning paths from the KGs for LLMs to reason. 
GoG~\cite{xu2024generate} improves upon ToG and allows more flexible traversing of KG instead of separate paths, 
and utilize LLM knowledge to answer questions if the LLM detects that the KG is incomplete. 
% Our GoG belongs to retrieval augmented methods, we also utilize the knowledge modeling ability of LLMs, as well as the semantic parsing ability and reasoning ability. 

% LLM reasoning with Prompting. 
% Many works have been proposed to elicit the reasoning ability of LLMs to solve complex tasks through prompting (Wei et al., 2023; Khot et al., 2023). Complex CoT (Fu et al., 2023) creates and refine rationale examples with more reasoning steps to elicit better reasoning in LLMs. Self-Consistency (Wang et al., 2023b) fully explores various ways of reasoning to improve their performance on reasoning tasks. De- comP (Khot et al., 2023) solves complex tasks by instead decomposing them into simpler sub-tasks and delegating these to sub-task specific LLMs. Re- Act (Yao et al., 2023) treats LLMs as agents that interact with the environment and make decisions to retrieve information from external source. GoG can be viewed as a fusion of ReAct and DecomP, thereby enabling a more comprehensive utilization of the diverse capabilities inherent in LLMs for addressing complex questions.

% While KGQA methods have strong relevance to RAG techniques, 
% as these methods also involve retrieving relevant information (triplets) from a large source. 
% Despite such strong relevance, however, 
While semi-structured can be regarded as a KG supplemented by text description on different entities, 
no existing KGQA methods consider how to utilize the additional text information, 
which can limits the answer quality if they are directly generalized to semi-structured data. 
Moreover, most entities in KG have a clear and unique name that can be used for reference, 
while such names do not always exist for semi-structured data 
and text information on each node can be important to distinguish different nodes. 
% as common knowledge graphs do not have additional text information for different entities. 

\begin{table*}
    % \resizebox{1.00\textwidth}{!}{
    \begin{tabular}{lccc}
        \toprule
         {Example question}  & Text information & Relational information \\
        \midrule
        \hline
       {\textit{What climbing guide do \textcolor{MidnightBlue}{most people purchase with}}} & {Black Diamond White } & {also\_buy} \\
        {\textit{\textcolor{OliveGreen}{Black Diamond White Gold Loose Chalk?}}}  & {Gold Loose Chalk} &  \\
        \midrule
        {\textit{What are some \textcolor{OliveGreen}{fashionable reversible bucket hats} }} & fashionable reversible bucket hats & text \\
        {\textit{that \textcolor{OliveGreen}{provide excellent sun protection against UVA and UVB rays}?}} &  & \\
        \midrule
        \midrule
         {\textit{Search {publications} \textcolor{MidnightBlue}{by Hao-Sheng Zeng} on \textcolor{OliveGreen}{non-Markovian dynamics}.}} & {non-Markovian dynamics}  & {author\_writes\_paper} \\
        % & \cellcolor{yellow!10}{} \\
        \midrule
        \textit{What are some \textcolor{OliveGreen}{nanofluid heat transfer} {research papers}} & \multirow{2}{*}{nanofluid heat transfer}  &  {author\_writes\_paper} \\
        \textit{published by scholars \textcolor{MidnightBlue}{from Philadelphia University}?} &  & {author\_affiliated\_with\_institution} \\
        \midrule
        \midrule
        \textit{What is the name of the condition characterized by } & a complete interruption & parent-child \\
        \textit{\textcolor{OliveGreen}{a complete interruption of the inferior vena cava},} & of the inferior vena cava \\
        \textit{ \textcolor{MidnightBlue}{falling under congenital vena cava anomalies}?}  \\
        \midrule
        % & \multirow{ 1}{*}{\texttt{disease}}
        % & \cellcolor{pink!20}\textit{Is my \textcolor{OliveGreen}{constant fatigue} due nto conditions \textcolor{MidnightBlue}{related to bartonellosis}?}\\
        \textit{Which \textcolor{OliveGreen}{gallbladder illness} serves as a \textcolor{MidnightBlue}{contraindication} to} & gallbladder illness, & contraindication \\
        \textit{\textcolor{MidnightBlue}{medications prescribed for} \textcolor{OliveGreen}{small cell neuroendocrine}} & small cell neuroendocrine & indication \\
        \textit{\textcolor{OliveGreen}{carcinoma of the urinary bladder}?} &  carcinoma of the urinary bladder & \\
        \midrule
        \textit{What are the pathways with both \textcolor{MidnightBlue}{a 'parent-child' hierarchy}} & TCR signaling pathway & parent-child \\ %  (4 more)
        \textit{related to the \textcolor{OliveGreen}{TCR signaling pathway}} & CD101 & interacts with \\ 
        \textit{and \textcolor{MidnightBlue}{interactions with} the \textcolor{OliveGreen}{CD101 gene or protein}?} \\ 
        \midrule
        \textit{What are the common \textcolor{MidnightBlue}{gene targets} for} & Hydrocortisone butyrate & target \\
        \textit{both \textcolor{OliveGreen}{Hydrocortisone butyrate} and \textcolor{OliveGreen}{2-Methoxyestradiol}?} & 2-Methoxyestradiol \\
        %& \cellcolor{pink!20}\textit{Which {gene} on \textcolor{OliveGreen}{chromosome 4} are \textcolor{MidnightBlue}{expressed in multicellular organisms}?} & {\small{<NFKB1>}},  {\small{<MT2P1>}}\cellcolor{pink!20}\\
        \bottomrule
    \end{tabular}
    % }
    \vspace{2pt}
    \caption{Example questions on semi-structured data. Relevant parts on \textcolor{MidnightBlue}{relational} and \textcolor{OliveGreen}{text} aspects are highlighted. 
    % Different background colors indicate queries on different data sets. 
    }
    \label{tab:benchmark_examples}
\end{table*}

\section{Problem Definition}
\label{sec:def}

We first give the formal definition of semi-structured data, 
which can be expressed by a text-attributed heterogeneous graph as follows:

\begin{definition}[Semi-structured data~\cite{wu2024stark}]
\label{def:semi}
Semi-structured data can be described by a text-attributed heterogeneous graph $\mathcal{G} = \{\mathcal{V}, \mathcal{E}, \mathcal{R}, f_{\mathcal{T}}, f_{\mathcal{R}}\}$, 
where $\mathcal{V}$ is the set of nodes, $\mathcal{E}$ is the set of edges,
% $\mathcal{N}$ is the set of node types, $\mathcal{R}$ is the set of edge types,
$f_{\mathcal{T}}: \mathcal{V} \rightarrow \text{text}$ 
is a mapping 
from nodes to their text description,
and $f_{\mathcal{R}}: \mathcal{E} \rightarrow \mathcal{R}$ is a mapping 
from edges
to edge types.
\end{definition}

An example of semi-structured data in biomedical domain is shown in Figure~\ref{fig:semi}. 
The nodes in $\gV$ are represented by blue boxes, 
and all texts in each blue box correspond to the text descriptions of each node obtained from $f_{\gT}$. 
Edge types from $f_{\gR}$ are shown near each edge in $\gE$. 
% These are all parts of the text description (i.e., the output of $f_{\mathcal{T}}$). 
With the above definition for semi-structured data, 
we can also formally define the QA task on such data as follows:

\begin{definition}[QA on semi-structured data]
\label{def:qa}
Given semi-structured data $\mathcal{G}$ defined in Definition~\ref{def:semi}, 
QA on this data set aims to find the answer nodes $a \in \mathcal{V}$ for a question $q$. 
% $\mathcal{G} = \{\mathcal{V}, \mathcal{E}, \mathcal{N}, \mathcal{R}, f_{\mathcal{N}}, f_{\mathcal{R}}\}$, 
% where $\mathcal{V}$ is the set of nodes, $\mathcal{E}$ is the set of edges,
% $\mathcal{N}$ is the set of node types, $\mathcal{R}$ is the set of edge types,
% $f_{\mathcal{N}}: \mathcal{V} \rightarrow \mathcal{N}$ 
% is a mapping 
% from nodes
% to node types,
% and $f_{\mathcal{R}}$ is a mapping 
% from edges
% to edge types.
\end{definition}

Note that our work. 
Some example questions on different semi-structured data are shown in Table~\ref{tab:benchmark_examples}. 
Compared with either standard RAG or KGQA task, 
retrieving contents from semi-structured data poses special challenges, 
as we need to jointly utilize relational and text information. 
% Here we provide some examples: 
% \begin{example}
%     Consider the question ``What drugs target the CYP3A4 enzyme and are used to treat strongyloidiasis?''. 
%     This question only involves relational information, 
%     as we can find the intersection set between drugs that target the CYP3A4 enzyme and treat strongyloidiasis. 
%     % Both relational information is present in. 
% \end{example}
% \begin{example}
%     Consider the question ``What drugs target the CYP3A4 enzyme and are used to treat strongyloidiasis?''. 
%     This question only involves relational information, 
%     as we can find the intersection set between drugs that target the CYP3A4 enzyme and treat strongyloidiasis. 
% \end{example}
We note that existing methods fail short to tackle such complex challenges. 
Standard RAG algorithms only use text information to directly answer the questions, 
and cannot perform multi-step reasoning based on relational information in semi-structured data. 
% but neglect relational information in graphs. 
On the contrary, existing KGQA methods neglect additional text information on different entities. 
Such limitations call for the development of novel methods 
that can jointly utilize relational and text information. 

% \begin{example}
%     Consider the question ``What drugs target the CYP3A4 enzyme and are used to treat strongyloidiasis?''. 
%     This question only involves relational information, 
%     as we can find the intersection set between drugs that target the CYP3A4 enzyme and treat strongyloidiasis. 
% \end{example}

\section{Proposed Method}
\begin{figure*}[h]
  \centering
  \includegraphics[width=0.95\textwidth]{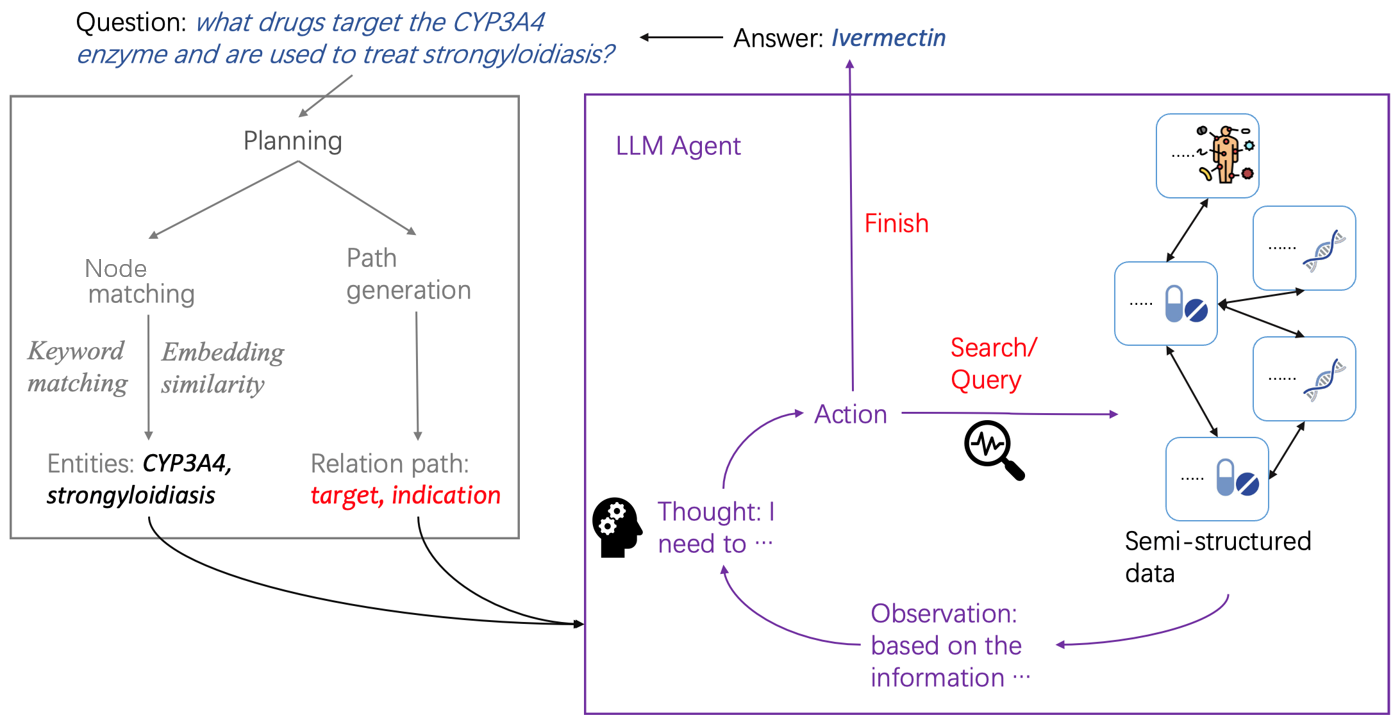}
  \vspace{-5px}
  \caption{Complete framework of PASemiQA consisting of two parts: planning module (section~\ref{ssec:plan}) and LLM agent (section~\ref{ssec:agent}).}
  \label{fig:met}
\end{figure*}

Following the limitation raised in section~\ref{sec:def}, 
in this section, we introduce how to effectively utilize the text and relational information together to answer questions with semi-structured data. 
The proposed method, called PASemiQA (Plan-Assisted Semi-structured data QA), 
is composed by two connected parts as in Figure~\ref{fig:met}. 
First, we introduce how to generate reasoning plans based on the given question,  
which consists of finding relevant nodes as well as edge types from the given semi-structured data, as in section~\ref{ssec:plan}. 
Then based on the generated plan, in section~\ref{ssec:agent} we propose to use a LLM model as an agent to traverse the whole data set and find the final answer. 
The complete algorithm is in section~\ref{ssec:alg}. 

\subsection{Planning Generation}
\label{ssec:plan}

% Recently, many techniques have been explored to improve the reasoning ability of LLMs by planning, which first prompts LLMs to generate a reasoning plan and then conduct reasoning based on it~\citep{wang2023plan}. 
% Similar ideas for planning have been proposed in KGQA~\cite{luo2024reasoning}.
% In this section, we extend such idea to answering questions with semi-structured data. 
% However, LLMs are known for having hallucination issues, which are prone to generating incorrect plans and leading to wrong answers \citep{ji2023survey}.

% To answer a given query defined in Definition~\ref{def:qa}, 
% we need to start from some entities, and find a path from these entities to the final answer. 
While answering a question defined in Definition~\ref{def:qa} may involve both text and relational information in the given semi-structured data, 
we propose to disentangle the complete reasoning process into the following two parts that separately depend on text and relational information:
\begin{itemize}
    \item Text information: given nodes with rich text description in the whole semi-structured data set, which nodes may be related to this question?
    \item Relational information: given numerous relations in the whole semi-structured data set, which relations may be related to this question?
\end{itemize}

To answer the two questions raised above, we introduce the following definition of generating a plan from the given question on semi-structured data:  

\begin{definition}[Planning generation on semi-structured data]
\label{def:plan}
With a semi-structured data set $\mathcal{G}$ defined in Definition~\ref{def:semi} and a question $q$ defined in Definition~\ref{def:qa}, 
planning generation maps this question $q$ to a set of nodes $\mathcal{V}_q \subseteq \mathcal{V}$, and a set of edge type sequences $\{ z_i \}$ 
where each $z_i = (r_{1, i}, \dots, r_{n_i, i})$ is a sequence of edge types (also called a {\bf relation path}) 
with each edge type $r_{j, i} \in \mathcal{R} \cup \{\text{text}\}, j=1, \dots, n_i$ and $n_i$ denoting the length of sequence $z_i$
\end{definition}

Note that the relation sequence consists of an additional ``type'' {\it text}, 
which indicates that the answer only depends on the text information in semi-structured data, 
and no relation in the given semi-structured data need to be involved to answer this question. 
Examples of queries from different semi-structured data sets are shown in Table~\ref{tab:benchmark_examples}, 
along with the extracted related text information from the question and edges in the semi-structured data. 
We then introduce how the plan generation module is designed for both text and relational information respectively. 
% An example utilize such relation is given below: 
% \begin{example}
% Consider the following question: ``What are some fashionable reversible bucket hats that provide excellent sun protection against UVA and UVB rays?''
% This query only involves searching for a ``fashionable reversible bucket hats'' that ``provide excellent sun protection against UVA and UVB rays''. 
% As such, the model should output ``VSS''.
% \end{example}
% Here is another example that does not involve VSS, 
% as the answer should be directly deduced by reasoning on the graph: 
% \begin{example}
% \label{exp2}
%     Consider the question ``What drugs target the CYP3A4 enzyme and are used to treat strongyloidiasis?''. 
%     This question only involves relational information, 
%     as we can find the intersection set between drugs that target the CYP3A4 enzyme and treat strongyloidiasis. 
%     The entity is CYP3A4 enzyme and strongyloidiasis.
%     Relation is target and treat. 
% \end{example}

\subsubsection{Text information}
For the first part on text information, 
a simple method utilized by most previous KGQA methods~\cite{sun2024thinkongraph, xu2024generate} is keyword matching, 
which matches names of nodes from the given KG in this question. 
The success of such method in KGQA methods is due to the clear reference of nodes by their names in KG, 
which may not still hold in the context of semi-structured data. 
As shown in the second row in Table~\ref{tab:benchmark_examples}, 
some queries does not directly point to a specific node, 
but gives node descriptions that cannot be well matched by exact name matching. 
Meanwhile, 
a standard baseline in various RAG methods~\cite{Zhao2024RetrievalAugmentedGF, wu2024stark} is based on the similarity between question and document embeddings, 
which in turn can be inaccurate when we need to match the exact names, 
such as the drug or gene names shown in Table~\ref{tab:benchmark_examples}. 
% there are naturally two types of entity matching methods: 
% \begin{itemize}
%     \item Keyword matching: matches exact keyword. While it may be accurate for biomedical data sets where different entities have exact names, 
%     it may not be so useful
%     \item Embedding similarity: select entities with largest embedding similarity. Compared with keyword matching it can be more flexible, but may fail short for more exact matching. 
% \end{itemize}

Motivated by the limitation of both methods, 
we then propose a hybrid solution to combine these two methods together. 
We first go through the question to extract all mentioned nodes based on their names as keyword (``Keyword matching'' in Figure~\ref{fig:met}). 
% The matching tries to find matching with maximum character. 
% For example, for the query ``
% For example, 
% consider the problem in Example~\ref{exp2}, 
% for two enzyme ``CYP'' and ``CYP3A4'', 
% while they are both matched in this query, 
% we only keep the latter one as it contains the maximum character. 
If we do not obtain any nodes in keyword matching, we will select top-$k$ nodes based on the cosine similarity between the text embedding for this question and the documents on different nodes (``Embedding similarity'' in Figure~\ref{fig:met}).
The matched nodes together form the node set $\mathcal{V}_q$ in Defition~\ref{def:plan}. 
% After the process is complete, 
% we obtain the entity set. 

% Consider this question: 
% Another question: What climbing guide do most people purchase with Black Diamond White Gold Loose Chalk?
% This involves the relationship “also\_buy” on the graph (the bolded part), and the model should output “also\_buy”.
% We fine-tuned a LLaMa2-7B model with SFT (Supervised Fine-Tuning) to accomplish the above task, and the two outputs were combined to form a complete search plan, which served as the basis for the subsequent search upstream of the graph.

% \RE{In this section, we introduce our method: reasoning on graphs (\ourmethod), which contains two components: 1) a \textit{planning} module that generates relation paths grounded by KGs as faithful plans; 2) a \textit{retrieval-reasoning} module that first retrieves valid reasoning paths from KGs according to the plans, then conducts faithful reasoning based on retrieved reasoning paths and generates answers with interpretable explanations. The overall framework of \ourmethod is illustrated in \Cref{fig:framework}.}

% \subsection{Reasoning on Graphs: Planning-Retrieval-Reasoning}

% %
% To address this issue, we present a novel \textit{planning-retrieval-reasoning} framework, which makes the reasoning plans grounded by KGs and then retrieves faithful reasoning paths for LLM reasoning. 
\subsubsection{Relational information}
% While the above procedure utilizes text information in semi-structured data, 
% another important factor is relational information. 
To utilize relational information in a given semi-structured data, 
as in Definition~\ref{def:plan},
we propose to generate a set of edge type sequences $\{ z_i \}$ 
that may be helpful to find the answers for a question. 
% how such information is utilized in query by generating relation paths
% to capture semantic relations between entities. 
% , have been utilized in many reasoning tasks on KGs \citep{wang2021relational,xusubgraph}. 
% Compared to the dynamically updated entities, the relations in KGs are more stable~\citep{wang2023survey}. 
% By using relation paths, we can always retrieve the latest knowledge from semi-structured data for reasoning. Therefore, relation paths can serve as faithful plans for reasoning. 
% By treating relation paths as plans, we can make sure the plans are grounded by KGs, which enables LLMs to conduct faithful and interpretable reasoning on graphs. In a nutshell, 
Specifically, given the question $q$ and semi-structured data $\gG$, 
the probability of choosing node $a$ as the answer can be expressed by conditioning on the relation path $z$
% we consider generation as the optimization problem that aims to maximize the probability of finding the correct answers from the semi-structured data $\gG$ w.r.t the query $q$ by expressing the probability of finding the answer entity $a$ 
as follows: 
% relation paths $z$ as follows:
% Thus, the objective function of our \ourmethod can be formulated as
\begin{equation}
    \label{eq:plan}
    P_\theta(a|q,\gG) = \sum_{z\in\gZ}P_{semi}(a|q,z,\gG)P_\theta(z|q),
\end{equation}
where $\gZ$ denotes the set of all possible relation paths. 
The first term $P_{semi}(a|q,z,\gG)$ refers to the probability of choosing node $a$ as the answer given the question $q$, relation path $z$ and semi-structured data $\gG$, 
and we let it to be uniform distribution on all nodes that can be reached by relation path $z$ from nodes in $\mathcal{V}_q$ obtained in previous section. 
% which naturally corresponds to entities that can be connected from relation path $z$ to an entity in the entity set $\mathcal{V}_q$. 
Mathematically, we can define $\mathcal{A}_{q,z}$ as the set of nodes $a$ that can be reached from any nodes $e \in \mathcal{V}_q$ as follows:
\begin{align*}
    \mathcal{A}_{q,z} = \{a:\text{there exists a relation path } z \text{ starts from } e \in \mathcal{V}_q \text{ to } a  \}
\end{align*}
% and can be mathematically expressed as follows:
% $\theta$ denotes the parameters of path generation model, $z$ denotes its generated relation paths, and 
% The first term $P(a|q,z,\gG)$ can take an easy form 
% that originates from the entity set $\mathcal{V}_q$ along the relation path $z$. 
Then $P_{semi}(a|q,z,\gG)$ takes the following form:
\begin{equation}
    \label{eq:semi}
    P_{semi}(a|q,z,\gG) =
    \left\{
    \begin{aligned}
         & \frac{1}{|\mathcal{A}_{q,z}|}, a \in \mathcal{A}_{q,z}, \\
         & 0, else,
    \end{aligned}
    \right.
\end{equation}
Now to maximize the probability of finding the correct answer in (\ref{eq:plan}), 
we need to optimize the latter term $P_\theta(z|q)$ 
that controls the generation of relation path $z$ given the question $q$. 
% Directly using it is not as the LLM cannot capture semantics. 
% Therefore, we propose to use instruction tuning. 
% Specifically, 
% The former term maximizes the expectation that retrieval-reasoning module generates correct answers based on the relation paths and KGs (retrieval-reasoning optimization). 
% \noindent\textbf{Planning optimization.} In planning optimization, we aim to distill the knowledge from KGs into LLMs to generate faithful relation paths as plans. 
% This can be achieved by minimizing the KL divergence with the posterior distribution of faithful relation paths $Q(z)$, 
% To generate faithful relation paths, 
% which can be approximated by the valid relation paths from the semi-structured data. 
Suppose we know the ground-truth answer $a^*$ for a question $q$, 
% Then given a question $q$ and possible answer entity $a$, 
then from a node $e \in \mathcal{V}_q$ obtained in previous section, 
if we can find the path instances $e\xrightarrow{r_1}\ldots\xrightarrow{r_l} a$ connecting these two nodes in semi-structured data, 
this relation path $z = \{ r_1, \dots,r_l\}$ should receive larger probability in $P_\theta(z|q)$. 
As such, we introduce the following distribution $Q(z|q, a^*, \gG)$ 
that $P_\theta(z|q)$ need to fit as follows: 
% The corresponding relation path $z=\{r_1,r_2,\dots,r_l\}$ can be considered valid and serve as a faithful plan for answering the question $q$. The posterior distribution $Q(z)$ can be formally approximated as
\begin{equation}
    \label{eq:posterior}
    % Q(z) \simeq Q(z|a, q, \gG)=
    Q(z|q, a^*, \gG) = 
    \left\{
    \begin{aligned}
         & \frac{1}{|\mathcal{Z}^*|}, z = (r_1, \dots, r_l), \exists  e \in \mathcal{V}_q, e\xrightarrow{r_1} \ldots\xrightarrow{r_l} a \in \gG  \\
         & 0, else,
    \end{aligned}
    \right.
\end{equation}
% where we assume a uniform distribution over all possible relation paths $\mathcal{Z}$, and $\exists w_z(e,a) \in \gG$ indicates the existence of a path instance connecting the entity $e \in \mathcal{V}_q$ and the ground-truth answer $a$ in $\gG$.
%
where we consider the shortest paths $\gZ^*\subseteq\mathcal{Z}$ between nodes $e \in \mathcal{V}_q$ and the ground-truth answer $a^*$. 
Then we define the following loss based on the KL divergence between the target distribution $Q(z|q, a^*, \gG)$ and the path generation model $P_\theta(z|q)$ as follows:
% \footnote{missing constant, please specify}
\begin{align}
    \gL_{\text{plan}} = & \KL(Q(z|q, a^*, \gG)\Vert P_\theta(z|q)) \notag \\
    = & - \frac{1}{|\gZ^*|}\sum_{z\in\gZ^*} \log P_\theta(z|q) + C,
    \label{eq:kl}
\end{align}
% The detailed derivation of can be found in \ref{app:planning_derivation}.
% By optimizing $\theta$ in (\ref{eq:kl}), we maximize the probability of path generation model to generate faithful relation paths from $\gG$, 
and the optimization of $\gL_{\text{plan}}$ can be achieved as
\begin{equation}
    \setlength\abovedisplayskip{1pt}%shrink space
    \setlength\belowdisplayskip{1pt}
    \argmax_\theta {\frac{1}{|\gZ^*|}}\sum_{z\in {\gZ^*}} \log P_\theta(z|q)= {\frac{1}{|\gZ^*|}} \sum_{z\in {\gZ^*}}\log \prod_{i=1}^{|z|}P_\theta(r_i|r_{<i},q),
\end{equation}
where 
% $P_\theta(z|q)$ denotes the prior distribution of generating faithful relation path $z$, and 
$P_\theta(r_i|r_{<i},q)$ denotes the probability of generating next relation $r_i$ for the relation path $z$ given all previous relations $r_{<i}$ and question $q$, 
and $|z|$ denotes the total length of the relation path $z$. 
In other words,  we maximize the probability of path generation model to generate faithful relation paths. 

For the implementation of $P_\theta(z|q)$, 
we finetune a pre-trained large language model (e.g, LlaMa model family~\cite{touvron2023llama}) by instruction fine-tuning. 
% we propose to utilize the instruction-following ability of LLMs \citep{wei2022finetuned}, and 
Specifically, we use the following template to construct the training corpus from any given semi-structured data set:
\begin{minipage}{0.95\columnwidth}
    \centering
    \vspace{5px}
    \begin{tcolorbox}
        % \small
        Please generate a valid relation path that can be helpful for answering the following question: \texttt{<Question>}. $z$
    \end{tcolorbox}
    \vspace{2px}
    % \vspace{1mm}
\end{minipage}

where \texttt{<Question>} will be replaced by the actual question $q$, 
followed by the ground-truth answer $z \in \gZ^*$ 
that is structurally formatted as a sentence like below:

\begin{minipage}{0.95\columnwidth}
    \centering
    \vspace{5px}
    \begin{tcolorbox}
        % \small
        $z=$ \texttt{<PATH>} $r_1$ \texttt{<SEP>}  $r_2$ \texttt{<SEP>} $\dots$ \texttt{<SEP>} $r_l$ \texttt{</PATH>}
    \end{tcolorbox}
    \vspace{2px}
\end{minipage}

where \texttt{<PATH>}, \texttt{<SEP>}, \texttt{</PATH>} are special tokens indicating the start, separator, and end of the relation path, respectively. 
Then at the inference stage, 
we will use the same prompt
% The question together with the instruction template is fed into LLMs 
to generate relation paths for the testing question $q$. 

\subsection{Graph Traversing Agent}
\label{ssec:agent}

% In this section, we introduce our method Generate-on-Graph (GoG), which can integrate the knowl- edge of KGs and LLMs, as well as utilize the rea- soning ability of LLMs. 
% The comparison between GoG and other previous methods is illustrated in Figure 3.

Based on the generated plan, we then need to actually retrieve relevant information from the semi-structured data. 
% we introduced the AGENT framework to break down answering the question into the following multi-round fixed process:
% \begin{itemize}
%     \item  Thought: based on the previous results to think about which action to take next (action)
%     \item Action: give the specific parameters of the next action, defined in detail as follows
%     \begin{itemize}
%         \item Search[entity]: query all neighboring nodes of the current entity on the map
%         \item Finish[entity]: reason the final answer based on the information obtained from the previous query.
%     \end{itemize}
%     \item Observation: process the information obtained on the graph as the basis for the next round of thought
% \end{itemize}
% LLM as Agent. 
While many recent works on KGQA~\cite{sun2024thinkongraph, xu2024generate} propose to use an LLM as an agent to perform step-by-step reasoning on KG, 
% we consider using an LLM as an agent that interacts with an environment (i.e., the semi-structured data) to solve tasks. 
these works cannot be directly generalized to semi-structured data as they only focus on relational information, 
and does not utilize additional text information in different nodes. 
As such, we propose to design a novel agent framework to answer questions on semi-structured data. 
This agent, implemented by an LLM model, 
first thinks of which action to take and express its {\it Thought} in natural languages, 
then performs the action to retrieve relevant information from the semi-structured data to obtain the set of triplets with node text description 
that will be used to determine next action. 
Such process is repeated for multiple times. 
The agent can take three different actions in total: {\sf Search}, {\sf Query} and {\sf Finish}, 
which focus on different aspects of semi-structured data:

% For semi-structured data, 
% As shown in Figure 3 (d), 
% Specifically, 
% we let the LLM agent perform a multi-round process to answer a given question. 
% We first generate think about which action to take next, 
% then give the specific parameters of the next action and process the information obtained on the graph as the basis for the next round of thought. 
% For the action, we consider three
% \begin{itemize}
%     \item  Thought: based on the previous results to think about which action to take next (action)
%     \item Action: give the specific parameters of the next action, defined in detail as follows
%     % \begin{itemize}
%     %     \item Search[entity]: query all neighboring nodes of the current entity on the map
%     %     \item Query[query]: use 
%     %     \item Finish[entity]: reason the final answer based on the information obtained from the previous query.
%     % \end{itemize}
%     \item Observation: process the information obtained on the graph as the basis for the next round of thought
% \end{itemize}
% For each step $i$, the agent first generates a thought $t_i$, where L is the language space, to decompose the original question (Thought 1) or decide which next sub-question should be solved (Thought 2). Then, based on the thought $t_i$, the agent chooses an action $a_i$ in the action space. 
\begin{enumerate}
    \item {\sf Search[node]}, which focuses on relational information from {\sf node}. 
    For simplicity, here we consider searching only one node, 
    and it is straight-forward to search multiple nodes by repeating the same process. 
    Given {\sf node}, 
    this action finds the most relevant top $K$ neighbors, 
    which uses the same strategy in ToG~\cite{sun2024thinkongraph} to filter unrelated neighbor nodes: first, we retrieve all neighbors that are connected to {\sf node}. 
% This process is completed by pre-defined SPARQL queries. 
Then, we utilize the LLM agent to select the top $K$ nodes that most related to current reasoning 
based on their edges to {\sf node}. 
%
% As there could be numerous neighbor entities of the target entity. 
% We  F
\item {\sf Query[query]}, which focuses on text information from {\sf query}. 
This action finds the most relevant top $K$ nodes for the given query 
based on the text description on different nodes. 
We utilize query embedding and the text embedding of each node's text description, 
and selects those nodes with top $K$ relevance scores computed by the cosine similarity between two embeddings.  
\item  {\sf Finish[answer]}, which indicates that the agent finishes the task with the node {\sf answer} as $a$ in Definition~\ref{def:qa}. 
% We also note the possibility that the agent generates ``Finish[unknown]'', which means that there is not enough information to answer the question while the agent does not know what to do next. 
% In such cases, we will roll back to the previous step and search one more hop neighbors of the last target entity to re-start the reasoning process.
\end{enumerate}
% Details are introduced below: 
% to search information by calling graph database API (Act 1, 2) or generate more information by reasoning and inherent knowledge (Act 4).

% Action Space. We use the selecting-generating-answering framework, which consists of two main actions: Search and Answer.

% As shown in Act 2 of Figure 3 (d), given the target entity Cupertino, LLMs select the two relation
% \{ (Cupertino, located\_in, California), (Cupertino, adjoin, Palo Alto) \} are appended to context as Obs 3.

% 2. Generate[sub-question], which make the agent generate new factual triples based on retrieval information and inherent knowledge. As shown in Act 4 of Figure 3 (d), although there is no triple directly representing the timezone of Cupertino, the agent can still infer the timezone of Cupertino based on {(Cupertino, located_in, California), (California, timezone, Pacific Standard Time)}. This process is similar to knowledge graph completion (KGC) (Wang et al., 2017). 

% It is also possible for the agent directly generates an entity, which is not explored before, based on its inherent knowledge. Therefore, we have to link the entity to its corresponding MID in the KG. This entity linking process is divided into two steps: (1) We retrieve some similar entities and their corresponding types based BM25 scores. (2) We utilize the LLM to select the most relevant
% entity based on the types.

% We will clarify these parts in revised version. 

The whole process will be repeated for $T$ rounds ($T$ is a hyper-parameter) to let the model terminate the traversing process on semi-structured data and return the final answer. 

\subsection{Complete Algorithm}
\label{ssec:alg}

The complete procedure of PASemiQA is shown in Algorithm~\ref{alg:semi}, 
with the exact prompt for the agent is elaborated in Appendix~\ref{app:prompt}. 
Compared to existing RAG or KGQA methods, 
PASemiQA jointly utilizes the text and relational information to answer the given question. 

\begin{algorithm}[ht]
	\caption{PASemiQA: Plan-Assisted Semi-structured data QA.}
	\begin{algorithmic}[1]
		\STATE Input: Node matching method $\text{match}$. Relation path generation model $P_{\theta}(z|q)$. LLM agent $agent$, question $q$; 
        % , observations $\gO$;
  \STATE Initialize the node set $\gE_0$ from the matching process: $\gE_0 = \{ e_i \} = match(q)$ and triplet set $\gO_0 = \{ \}$ to empty
  \STATE Generate relation paths $\{ z_j \}$ from the path generation model $P_{\theta}(z|q)$
		\FOR{$t = 0, \dots, T-1$}
		\STATE Generate thought $t$ from previous triplet set $\gO_t$
  \STATE Generate {\sf Action} based on thought $t$. 
  \IF{{\sf Action} is {\sf Finish[answer]}}
  \STATE Break
  \ELSIF{{\sf Action} is {\sf Query[query]}}
  \STATE Return $k$ nodes with largest similarity score between their document embedding and {\sf query} to form the node set $\gE_t$
  \ELSIF{{\sf Action} is {\sf Search[query]}}
  \STATE Retrieve triplets $(e_t, r, e_{t+1})$ from semi-structured data $\gG$ based on existing nodes $e_t \in \gE_t$ and relation paths $\{ z_j \}$
  \STATE Append new node $e_{t+1}$ to the node set: $\gE_{t+1} = \gE_t \cup \{ e_{t+1} \}$
  \ENDIF
  \STATE Append retrieved information to triplet set: $\gO_{t+1} = \gO_t \cup \{ (e_t, r, e_{t+1}) \}$
		\ENDFOR
  \STATE Output the {\sf answer} from the last {\sf Finish} action
	\end{algorithmic}
	\label{alg:semi}
\end{algorithm}

\section{Experiments}

We conduct experiments on three semi-structured data sets from the STaRK benchmark~\cite{wu2024stark} that focus on diverse fields: 
Amazon (product recommendation), MAG (academic network) and PrimeKG (biomedical network). 
The data split follows~\cite{wu2024stark}, 
and detailed information and statistics of these data sets are in Table~\ref{tab:statistics} in Appendix~\ref{app:data}. 
For all experiments, unless otherwise specified, 
we 
finetune a LLaMa2-7B model~\cite{touvron2023llama} as the path generation model in section~\ref{ssec:plan}, 
and 
use the GPT-4 model as the LLM agent in section~\ref{ssec:agent}. 
Following standard practice in KGQA methods~\cite{xu2024generate, luo2024reasoning, sun2024thinkongraph}, 
we use the Hit@1 score as the performance metric for different methods.

We compare the proposed method against baseline methods both from RAG on semi-structured data and KGQA methods. 
For RAG methods, we choose two baseline methods in~\cite{wu2024stark}: 
VSS and VSS+GPT-4 reranker. 
We use VSS as a representative for RAG methods as it is simple, fast and easy to implement, 
and VSS+GPT-4 reranker achieves the best performance among all methods in~\cite{wu2024stark}. 
For KGQA methods, we choose ToG~\cite{sun2024thinkongraph}, RoG~\cite{luo2024reasoning} 
and GoG~\cite{xu2024generate} 
as recent representatives, 
which achieves satisfying performance on existing KGQA data sets such as
CWQ~\cite{talmor2018web} or WebQSP~\cite{yih2016value}. 
For fair comparison, all KGQA methods also use the GPT-4 model for their agents. 

\begin{table}[h]
  \caption{Hit@1 scores on different semi-structured data with different methods. Best highlighted in bold.}
  \label{tab:exp}
  \vspace{-5px}
  \begin{tabular}{c c c c c}
    \toprule
    % Method & MRR & Hits@1 & Hits@3 & Hits@10 \\
    % \midrule
    % RotatE & 0.279 & 0.184 & 0.248 & 0.432 \\
    % NBFNet & 0.177 & 0.112 & 0.168 & 0.296 \\
    dataset	& PrimeKG & MAG & Amazon \\
    \midrule
    VSS & 0.1263 & 0.2908 & 0.4044 \\
    VSS+GPT-4 reranker & 0.1828 & 0.4090 & 0.4479 \\
    \midrule
    ToG & 0.1321 & 0.0540 & 0.1667 \\
    RoG & 0.2274 & 0.3294 & 0.3122 \\
    GoG & 0.1892 & 0.4126 & 0.2101 \\
    \midrule
    PASemiQA (GPT-4 Agent) & {\bf 0.2968} & {\bf 0.4316} & {\bf 0.4586} \\
    \bottomrule
    \end{tabular}
\end{table}

\begin{table}[h]
  \caption{Wall-clock time cost (mean$\pm$std in seconds) of different methods to answer a question with given semi-structured data.}
  \label{tab:latency}
  \vspace{-5px}
  \begin{tabular}{c c c c c}
    \toprule
    % Method & MRR & Hits@1 & Hits@3 & Hits@10 \\
    % \midrule
    % RotatE & 0.279 & 0.184 & 0.248 & 0.432 \\
    % NBFNet & 0.177 & 0.112 & 0.168 & 0.296 \\
    	& PrimeKG & MAG & Amazon \\
    \midrule
    VSS & 0.54$\pm$0.01 & 2.25$\pm$0.01 & 5.71$\pm$0.02 \\
    VSS+GPT-4 reranker & 26.97$\pm$1.96 & 23.43$\pm$1.64 & 24.76$\pm$1.55 \\
    \midrule
    ToG & 14.68$\pm$1.57 & 24.16$\pm$1.53 & 32.26$\pm$1.74 \\
    RoG & 7.89$\pm$0.98 & 11.76$\pm$1.07 & 10.42$\pm$1.02 \\
    GoG & 28.56$\pm$2.02 & 24.73$\pm$1.79 & 32.68$\pm$1.65 \\
    \midrule
    PASemiQA (GPT-4 Agent) & 28.19$\pm$2.04 & 25.48$\pm$1.77 & 18.74$\pm$1.34 \\
    \bottomrule
    \end{tabular}
\end{table}

\subsection{Comparison on Accuracy and Latency}
% \begin{itemize}
%     \item RAG
%     \item KGQA: ToG, RoG, GoG
% \end{itemize}

Table~\ref{tab:exp} compares the answering accuracy of different methods for questions on semi-structured data. 
Due to the space limit, results with other performance metrics (e.g., MRR or macro F1 scores) are in Appendix~\ref{app:exp}.  
% Despite its simplicity, VSS achieves the worst overall performance, 
% which demonstrates that solely utilizing text information from embedding similarity is not effective to derive correct answers. 
For RAG methods, 
VSS+GPT-4 reranker almost outperforms all the other baseline on MAG and Amazon data set. 
% We can first see a clear gap between different data sets. 
For Amazon data set where questions solely depending on text information make a large proportion, 
RAG baselines (VSS and VSS+GPT-4 reranker) achieves much higher performance than KGQA methods that cannot effectively utilize text information. 
For PrimeKG data set where some questions require complex relational information, 
KGQA methods significantly outperforms RAG baselines. 
The proposed method achieves the best overall performance, 
which demonstrate the necessity of joint utilization of both text information and relational information 
to answer questions on semi-structured data. 

We further compare the time cost of different methods in Table~\ref{tab:latency}. 
Both VSS and RoG has much smaller time cost to answer a given question than other methods, 
as they do not need to iteratively select relational information from semi-structured data. 
ToG also has much smaller time cost on PrimeKG data set, 
yet its performance on this data set is also much worse than other KGQA methods, 
which may come from ineffective utilization of the relational information 
as this data set is highly professional and contains biomedical terms that may not be easy to understand. 
We can see that the proposed method achieves comparable time cost with VSS with reranker methods,
but achieve better performance. 
In other words, 
the proposed method better utilizes the language understanding abilities of LLMs as an agent 
to find correct answers for a given question. 

% \begin{table*}[t]
%     \centering
%     \caption{Latency (s) of different retrieval systems. }
%     \resizebox{1.0\textwidth}{!}{
%     \begin{tabular}{l|cccccc}
%         \toprule
%         &\textbf{\small Dense Retriever} & \textbf{\small QAGNN } & 
%         \textbf{\small VSS} & \textbf{\small Multi-VSS} & \textbf{\small VSS+Claude} & \textbf{\small VSS+GPT4} \\
%         \midrule
%         \textbf{Amazon} & 2.34 & 2.32 &5.71 &4.87 & 27.24&24.76 \\
%         \textbf{MAG} & 0.94 &1.35 &2.25 &3.14 &22.60 & 23.43 \\
%         \textbf{PrimeKG} & 0.92 & 1.29 &0.54 &0.90 &29.14 &26.97 \\
%         \midrule 
%         \textbf{Average} & 1.40  & 1.65 &2.83 &2.97 &26.33 &25.05 \\
%         \bottomrule
%     \end{tabular}
%     }
%     \label{tab:latency}
% \end{table*}

\begin{table}[h]
  \caption{Hit@1 scores of PASemiQA with different LLM agents on semi-structured data. Best highlighted in bold.}
  \label{tab:llm}
  \vspace{-5px}
  \begin{tabular}{c c c c c}
    \toprule
    % Method & MRR & Hits@1 & Hits@3 & Hits@10 \\
    % \midrule
    % RotatE & 0.279 & 0.184 & 0.248 & 0.432 \\
    % NBFNet & 0.177 & 0.112 & 0.168 & 0.296 \\
    	& PrimeKG & MAG & Amazon \\
    \midrule
    Qwen2-72B & 0.1849 & 0.0976 & 0.4163 \\
    LlaMa3-70B & 0.2392 & 0.4113 & 0.3264 \\
    GPT-3.5 & 0.2456 & 0.3092 & 0.3781 \\
    GPT-4 & {\bf 0.2968} & {\bf 0.4316} & {\bf 0.4586} \\
    \bottomrule
    \end{tabular}
\end{table}

\begin{table}[h]
  \caption{Hit@1 scores on different semi-structured data with different node matching methods. Best highlighted in bold.}
  \label{tab:match}
  \vspace{-5px}
  \begin{tabular}{c c c c c}
    \toprule
    % Method & MRR & Hits@1 & Hits@3 & Hits@10 \\
    % \midrule
    % RotatE & 0.279 & 0.184 & 0.248 & 0.432 \\
    % NBFNet & 0.177 & 0.112 & 0.168 & 0.296 \\
    dataset	& PrimeKG & MAG & Amazon \\
    \midrule
    % Qwen2-72B & 0.18 & 0.0976 & 0.41\\
    % LlaMa3-70B & 0.2392 & 0.4113 & 0.32 \\
    % GPT-3.5 & 0.2456 & 0.3092 & 0.37 \\
    Exact name matching & 0.2945 & 0.4062 & 0.2102 \\
    Embedding similarity & 0.2598 & 0.4158 & 0.3978 \\
    PASemiQA & {\bf 0.2968} & {\bf 0.4316} & {\bf 0.4586} \\
    \bottomrule
    \end{tabular}
\end{table}

\begin{table}[h]
  \caption{Hit@1 scores on different semi-structured data with different plan generation modules. Best highlighted in bold.}
  \label{tab:rel}
  \vspace{-5px}
  \begin{tabular}{c c c c c}
    \toprule
    % Method & MRR & Hits@1 & Hits@3 & Hits@10 \\
    % \midrule
    % RotatE & 0.279 & 0.184 & 0.248 & 0.432 \\
    % NBFNet & 0.177 & 0.112 & 0.168 & 0.296 \\
    	& PrimeKG & MAG & Amazon \\
    \midrule
    % Qwen2-72B & 0.18 & 0.0976 & 0.41\\
    % LlaMa3-70B & 0.2392 & 0.4113 & 0.32 \\
    % GPT-3.5 & 0.2456 & 0.3092 & 0.37 \\
    None & 0.2868 & 0.4068 & 0.3754 \\
    GPT-4 in-context & 0.2674 & 0.4148 & 0.4162 \\
    PASemiQA & {\bf 0.2968} & {\bf 0.4316} & {\bf 0.4586} \\
    \bottomrule
    \end{tabular}
\end{table}

% \begin{table}[ht]
%   \caption{Hit@1 scores on different semi-structured data with different values of $K$. $T$ is set to 5 by default.}
%   \label{tab:numk}
%   \vspace{-10px}
%   \begin{tabular}{c c c c c}
%     \toprule
%     % Method & MRR & Hits@1 & Hits@3 & Hits@10 \\
%     % \midrule
%     % RotatE & 0.279 & 0.184 & 0.248 & 0.432 \\
%     % NBFNet & 0.177 & 0.112 & 0.168 & 0.296 \\
%    	& PrimeKG & MAG & Amazon \\
%     \midrule
%     % Qwen2-72B & 0.18 & 0.0976 & 0.41\\
%     % LlaMa3-70B & 0.2392 & 0.4113 & 0.32 \\
%     % GPT-3.5 & 0.2456 & 0.3092 & 0.37 \\
%     $K=1$ & 0.2772 & 0.4166 & 0.3989 \\
%     $K=3$ & 0.2945 & 0.4289 & 0.4537 \\
%     $K=5$ & 0.2968 & 0.4316 & 0.4586 \\
%     $K=7$ & 0.2957 & 0.4298 & 0.4564 \\
%     \bottomrule
%     \end{tabular}
% \end{table}

% \begin{table}[h]
%   \caption{Wall-clock time cost (s) of PASemiQA with different values of $K$ to answer a question. $T$ is set to 5 by default.}
%   \label{tab:latk}
%   \vspace{-10px}
%   \begin{tabular}{c c c c c}
%     \toprule
%     % Method & MRR & Hits@1 & Hits@3 & Hits@10 \\
%     % \midrule
%     % RotatE & 0.279 & 0.184 & 0.248 & 0.432 \\
%     % NBFNet & 0.177 & 0.112 & 0.168 & 0.296 \\
%     & PrimeKG & MAG & Amazon \\
%     \midrule
%     $K=1$ & 24.37 & 18.82 & 11.98 \\
%     $K=3$ & 26.65 & 22.14 & 15.06 \\
%     $K=5$ & 28.19 & 25.48 & 18.74 \\
%     $K=7$ & 29.24 & 28.26 & 22.36 \\
%     \bottomrule
%     \end{tabular}
% \end{table}

\begin{figure}[ht]
  \centering
  \subfigure[Hit@1 score. \label{fig:numk}]{{\includegraphics[width=0.23\textwidth]{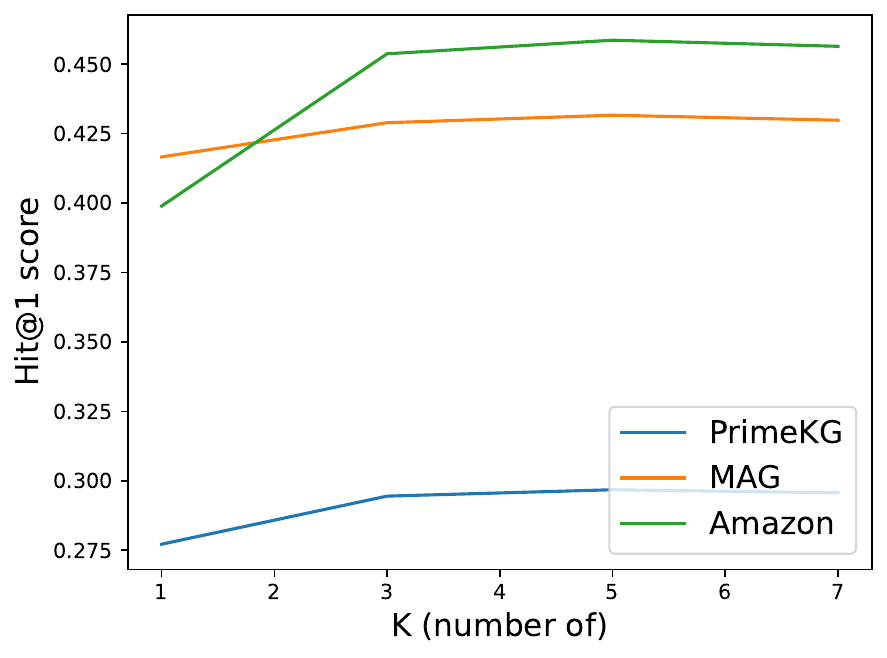}}}
  \subfigure[Per-question answering time. \label{fig:latk}]{{\includegraphics[width=0.23\textwidth]{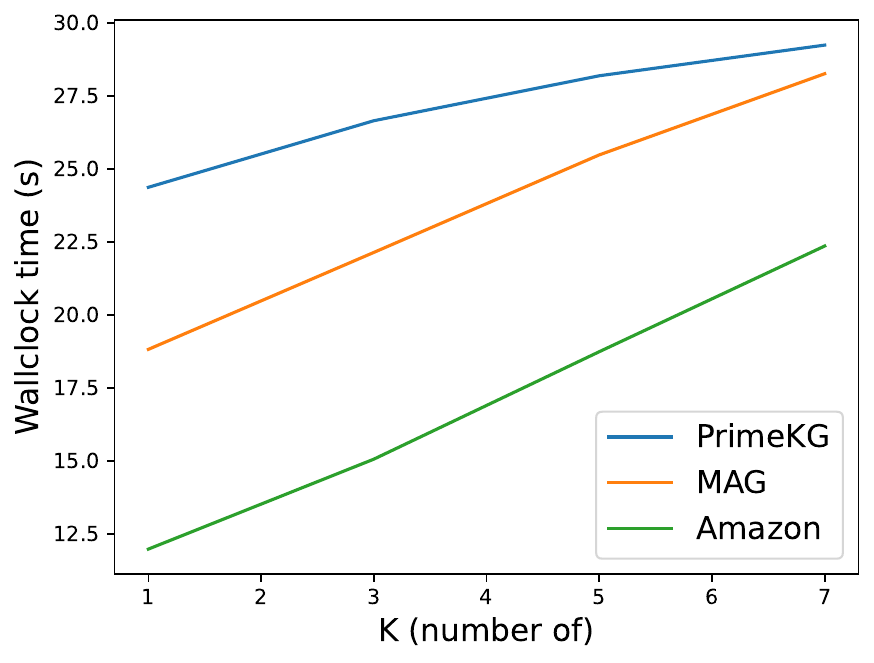}}}
  \vspace{-5px}
  \caption{Comparison of PASemiQA with different values of $K$. $T$ is set to 5 by default.}
  \label{fig:k}
\end{figure}

% \begin{table}[ht]
%   \caption{Hit@1 scores on different semi-structured data with different values of $T$. $K$ is set to 5 by default.}
%   \label{tab:numt}
%   \vspace{-10px}
%   \begin{tabular}{c c c c c}
%     \toprule
%     % Method & MRR & Hits@1 & Hits@3 & Hits@10 \\
%     % \midrule
%     % RotatE & 0.279 & 0.184 & 0.248 & 0.432 \\
%     % NBFNet & 0.177 & 0.112 & 0.168 & 0.296 \\
%     & PrimeKG & MAG & Amazon \\
%     \midrule
%     % Qwen2-72B & 0.18 & 0.0976 & 0.41\\
%     % LlaMa3-70B & 0.2392 & 0.4113 & 0.32 \\
%     % GPT-3.5 & 0.2456 & 0.3092 & 0.37 \\
%     $T=1$ & 0.2772 & 0.4166 & 0.3989 \\
%     $T=3$ & 0.2945 & 0.4289 & 0.4537 \\
%     $T=5$ & 0.2968 & 0.4316 & 0.4586 \\
%     % $K=7$ & 0.2957 & 0.4298 & 0.4564 \\
%     \bottomrule
%     \end{tabular}
% \end{table}

% \begin{table}[h]
%   \caption{Wall-clock time cost (s) of PASemiQA with different values of $T$ to answer a question. $K$ is set to 5 by default.}
%   \label{tab:latt}
%   \vspace{-10px}
%   \begin{tabular}{c c c c c}
%     \toprule
%     % Method & MRR & Hits@1 & Hits@3 & Hits@10 \\
%     % \midrule
%     % RotatE & 0.279 & 0.184 & 0.248 & 0.432 \\
%     % NBFNet & 0.177 & 0.112 & 0.168 & 0.296 \\
%     	& PrimeKG & MAG & Amazon \\
%     \midrule
%     $T=1$ & 16.72 & 13.65 & 7.88 \\
%     $T=3$ & 24.26 & 19.72 & 14.03 \\
%     $T=5$ & 28.19 & 25.48 & 18.74 \\
%     \bottomrule
%     \end{tabular}
% \end{table}

\begin{figure}[ht]
  \centering
  \subfigure[Hit@1 score. \label{fig:numt}]{{\includegraphics[width=0.23\textwidth]{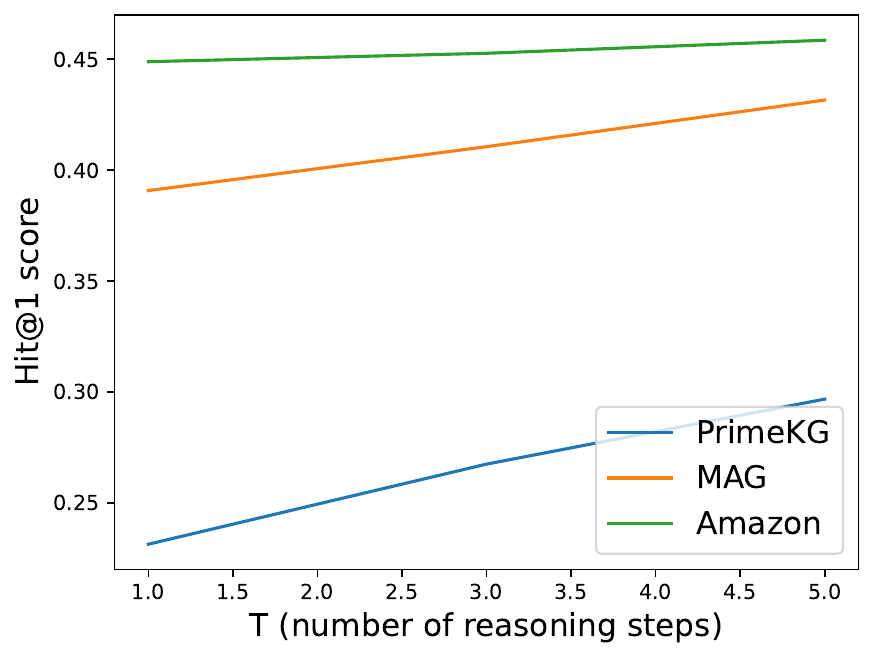}}}
  \subfigure[Per-question answering time. \label{fig:latt}]{{\includegraphics[width=0.23\textwidth]{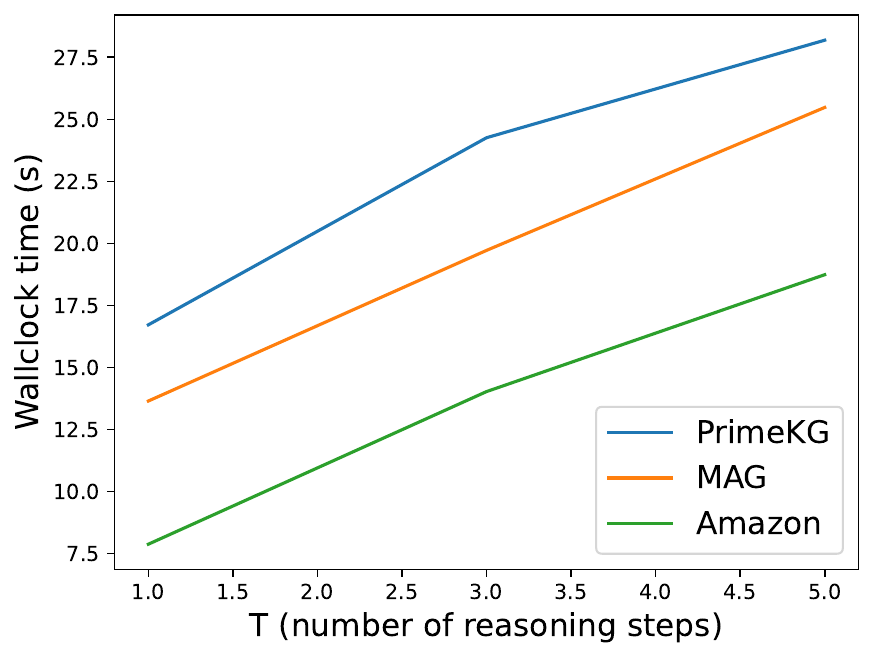}}}
  \vspace{-5px}
  \caption{Comparison of PASemiQA with different values of $T$. $K$ is set to 5 by default.}
  \label{fig:t}
\end{figure}

\subsection{Ablation Studies}

\subsubsection{Using Different LLM Models as the Agent}

While our proposed method does not limit the choice of LLM model as the agent, 
we compare the performance of different LLM models. 
We consider two open-source models: LLaMa 3-70B~\cite{dubey2024llama} and Qwen2-72B~\cite{yang2024qwen2}, 
as well as OpenAI's GPT-3.5~\cite{brown2020language} and GPT-4~\cite{achiam2023gpt} models as two representative closed-source LLMs. 
The other closed-source LLMs (e.g., Gemini~\cite{team2023gemini} or Claude~\cite{caruccio2024claude}) are not compared due to the additional API costs. 

Table~\ref{tab:llm} compares the performance of these models when used as the agent. 
We can see that GPT-4 model generally performs the best, 
while GPT-3.5 model performs worse than GPT-4 on all three data sets. 
Two open-source models achieve exceptional performance on specific semi-structured data: 
Qwen2 model only slightly falls short from GPT-4 on Amazon data set, 
while LLaMa3 model achieves almost comparable performance with GPT-4 on MAG data set. 
One possible reason for the performance difference across these semi-structured data 
is the differences in the pre-training corpus for these models, 
which may also motivate future works training an LLM agent specifically target on the given semi-structured data. 
% Compared to the more advanced model GPT-4, 
% GPT-3.5 model performs slightly worse on all three data sets, 
% and does not have. 

\begin{table*}[ht]
  \caption{Example planning and reasoning process of PASemiQA on a question from the MAG data set.}
  \vspace{-5px}
  \label{tab:case2}
  \begin{tabular}{c | c}
    \toprule
    Question $q$ & Show me publications by Th. Friedrich on high heat flux experiments with divertor components. \\
    \midrule
    Topic Node $\gV_q$ &  [Th. Friedrich] \\
    \midrule
    Plan $\{z_i\}$ & author\_\_\_writes\_\_\_paper $z_1$ \\
    \midrule
    \midrule
    Thought 1 & I need to search for publications by Th. Friedrich \\
    & specifically focusing on high heat flux experiments with divertor components. \\
    \midrule
    Action 1 & Search[Th. Friedrich] \\
    \midrule
    Triplet Set 1 & Th. Friedrich ($e_0$), author\_\_\_writes\_\_\_paper ($r_0$), \\
    $\gO_1$ & Design improvement of the target elements of Wendelstein 7-X divertor ($e_1$) \\
    & Th. Friedrich ($e_0$), author\_\_\_writes\_\_\_paper ($r_0$), \\
    & Results and consequences of high heat flux testing as quality assessment of the Wendelstein 7-X divertor ($e_2$) \\
\midrule
\midrule
Thought 2 & "Results and consequences of high heat flux testing as quality assessment of the Wendelstein 7-X divertor" \\
& is a publication by Th. Friedrich that focuses on high heat flux experiments with divertor components.", \\
\midrule
Action 2 & Finish[Results and consequences of high heat flux testing as quality assessment of the Wendelstein 7-X divertor] \\
& (Finish with answer $a$) \\
    \bottomrule
    \end{tabular}
    \vspace{-10px}
\end{table*}

\subsubsection{Implementation of Planning Modules}

The planning module in section~\ref{ssec:plan} involves two parts: 
node matching to utilize text information and plan generation to utilize relational information. 
In this section, we investigate how different design choices of these two parts may impact the performance of the proposed framework.

For node matching, we compare the proposed hybrid matching method in PASemiQA with solely using node matching or embedding similarity, 
and the performance comparison on different semi-structured data sets is shown in Table~\ref{tab:match}. 
We can see that these methods have different performances on different data sets. 
Node matching achieves satisfying performance for PrimeKG as biomedical objects have unique names and can be exactly matched. 
Meanwhile, for Amazon data set, 
the item names are often not stable and we need to better utilize the text information from each node. 
The performance for MAG is generally stable across different methods. 
Across different data sets, the proposed method achieves the best overall performance 
due to its flexibility on different semi-structured data. 

For planning generation module, 
we compare against the following baseline: 
(i) None, 
which directly uses the LLM agent from the node set $\mathcal{V}_q$ 
to traverse the semi-structured data and find the answer, 
(ii) GPT-4 in-context, 
% Despite fine-tuning a LLM model to generate, 
% other baseline also exists for plan generation. 
% Another baseline is to 
which utilizes the instruction following and few-shot learning ability of existing LLM models 
to generate the relation path without fine-tuning. 
Similar to the prompt in section~\ref{ssec:plan}, 
the in-context learning prompt for path generation is designed as follows:

\begin{minipage}{0.95\columnwidth}
    \centering
    \vspace{5px}
    \begin{tcolorbox}
        % \small
        Please generate a valid relation path that can be helpful for answering the following question. Examples are listed below:

        \texttt{<Question 1> $q_1$}: \texttt{<PATH>} $r_{1,1}$ \texttt{<SEP>}  $r_{1,2}$ \texttt{<SEP>} $\dots$ \texttt{<SEP>} $r_{1,l_1}$ \texttt{</PATH>}

        ...

        \texttt{<Question n> $q_n$}: \texttt{<PATH>} $r_{n,1}$ \texttt{<SEP>}  $r_{n,2}$ \texttt{<SEP>} $\dots$ \texttt{<SEP>} $r_{n, l_n}$ \texttt{</PATH>}
        
        \texttt{<Question> $q$}: 
    \end{tcolorbox}
    \vspace{2px}
    % \vspace{1mm}
\end{minipage}

Compared to the prompt in section~\ref{ssec:plan}, 
we add some additional examples from the given semi-structured data for in-context learning on the semantics of different types of edges. 

Table~\ref{tab:rel} compares the performance with different path generation methods. 
We can see that even without the relation path generation module, 
the proposed method still achieves better performance than other baseline methods on PrimeKG data set, 
and the performance is comparable in MAG data set. 
However, for Amazon data set, the proposed method cannot even outperforms most simple VSS baseline without relation path generation. 
This may partially be attributed to large proportion of questions that only requires text information in Amazon data set. 
Solely utilizing relational information is not helpful for answering these queries, 
and we need to use a path generation module to determine whether text information is enough to answer the given question. 
% The proposed method achieves the best overall performance. 
We can also see that the influence of path generation module is more significant for Amazon and PrimeKG data set, 
but becomes less significant for MAG data set. 
While we have explained the influence on Amazon data set from the large proportion of questions that only depend on text information, 
the reason on PrimeKG data set is due to its complex types of edges from the biomedical domain, 
which may not be so familiar for general LLM models.

\subsection{Hyper-parameter Sensitivity}

In Algorithm~\ref{alg:semi}, the proposed method PASemiQA involves two hyper-parameters $K$ and $T$. 
$K$ controls the number of related nodes extracted in each step, 
while $T$ controls the total number of steps executed to answer each question. 
These two hyper-parameters can impact both the answer accuracy and latency of the proposed method, 
which will be analyzed in this section. 

Figure~\ref{fig:numk} compares the performance with different values of $K$, 
and 
Figure~\ref{fig:latk} compares the time cost with different values of $K$. 
% Throughout the experiments we use the GPT-4 model. 
Intuitively, setting $K$ either too large or too small will cause performance downgradation: 
small $K$ cannot ensure sufficient exploration in the whole semi-structured data, 
while large $K$ may include too much unrelated information in our reasoning process. 
Moreover, setting $K$ too large may also lead to increasing time cost, 
as the agent needs to process more nodes. 
The empirical results generally agrees with such intuition that 
either setting $K$ too small or too large leads to sub-optimal performance. 
We also note the performance difference is not significant across different values of $K$, 
which indicates that the performance of proposed method is robust to this hyper-parameter. 

Figure~\ref{fig:numt} compares the performance with different values of $T$, and 
Figure~\ref{fig:latt} compares the time cost with different values of $T$. 
Intuitively, we need to set $T$ large enough so as to ensure the complete semi-structured data are explored. 
Nevertheless, setting $T$ too large may also lead to increasing latency, 
as the agent needs to repeat the reasoning process for more times, 
even it can control the number of steps with ``Finish'' action. 
The empirical results generally agrees with such intuition that 
either setting $T$ too small or too large leads to sub-optimal performance. 
We also note the performance difference is not significant across different values of $T$, 
which indicates that the performance of proposed method is robust to this hyper-parameter. 

\subsection{Case Studies}

% In this section, we us to analyze how our method jointly uses text and relational information to answer questions. 
% Table~\ref{tab:case1} shows the reasoning process of a question. 
% This question solely depends on traversing relational information in semi-structured data. 
% We start with topic entities, 
% and find related entities. 
% While the answer is already present in the first step, 
% the agent needs to check again to ensure that, 
% as no information is directly given from CYP3A4 enzyme. 
% After that, the agent selects Ivermectin instead of Thiabendazole 
% based on its connection to CYP3A4 enzyme, 
% and finish the reasoning process. 
Finally, we use a real example in Table~\ref{tab:case2} to demonstrate how the proposed method PASemiQA jointly uses text and relational information to answer questions. 
% Table~\ref{tab:case2} shows the reasoning process of answering a question with the MAG data set. 
% This question solely depends on traversing relational information in semi-structured data. 
With its planning module, 
PASemiQA first identifies the topic node to be the author ``Th. Friedrich'', 
and the generated relation path contains only one relation ``author\_\_\_writes\_\_\_paper'' 
as this question is about the author's publication. 
Then starting with the topic node, 
PASemiQA lets the LLM agent to use the generated relation path 
to find more nodes related to the question. 
% While the answer is already present in the first step, 
% the agent needs to check again to ensure that, 
% as no information is directly given from CYP3A4 enzyme. 
% After that, the agent selects Ivermectin instead of Thiabendazole 
% based on its connection to CYP3A4 enzyme, 
% After the first step is completed, 
The LLM agent further uses its own language understanding ability 
on the paper title to select the correct answer
and finish the reasoning process.

\section{Conclusion}

In this paper, we design a novel method PASemiQA to jointly utilize text and relational information for answering questions with semi-structured data. 
PASemiQA first obtains a plan to determine text and relational information in semi-structured data
that is useful to answer the given question. 
Then based on the generated plan, 
PASemiQA introduces an agent implemented by an LLM to traverse the semi-structured data and
extract related information for this question. 
% extracts related entities from the given query 
% based on both keywords and text embeddings. 
Empirical results across semi-structured data sets from different domains demonstrate the effectiveness of the proposed method. 

% \begin{acks}
% \end{acks}

\bibliographystyle{ACM-Reference-Format}
\bibliography{sample-base}


\begin{thebibliography}{31}

%%% the Web address.

\ifx \showCODEN    \undefined \def \showCODEN     #1{\unskip}     \fi
\ifx \showDOI      \undefined \def \showDOI       #1{#1}\fi
\ifx \showISBNx    \undefined \def \showISBNx     #1{\unskip}     \fi
\ifx \showISBNxiii \undefined \def \showISBNxiii  #1{\unskip}     \fi
\ifx \showISSN     \undefined \def \showISSN      #1{\unskip}     \fi
\ifx \showLCCN     \undefined \def \showLCCN      #1{\unskip}     \fi
\ifx \shownote     \undefined \def \shownote      #1{#1}          \fi
\ifx \showarticletitle \undefined \def \showarticletitle #1{#1}   \fi
\ifx \showURL      \undefined \def \showURL       {\relax}        \fi
% invisible to TeX
\providecommand\bibfield[2]{#2}
\providecommand\bibinfo[2]{#2}
\providecommand\natexlab[1]{#1}
\providecommand\showeprint[2][]{arXiv:#2}

\bibitem[Achiam et~al\mbox{.}(2023)]%
        {achiam2023gpt}
\bibfield{author}{\bibinfo{person}{Josh Achiam}, \bibinfo{person}{Steven
  Adler}, \bibinfo{person}{Sandhini Agarwal}, \bibinfo{person}{Lama Ahmad},
  \bibinfo{person}{Ilge Akkaya}, \bibinfo{person}{Florencia~Leoni Aleman},
  \bibinfo{person}{Diogo Almeida}, \bibinfo{person}{Janko Altenschmidt},
  \bibinfo{person}{Sam Altman}, \bibinfo{person}{Shyamal Anadkat},
  {et~al\mbox{.}}} \bibinfo{year}{2023}\natexlab{}.
\newblock \showarticletitle{Gpt-4 technical report}.
\newblock \bibinfo{journal}{\emph{arXiv preprint arXiv:2303.08774}}
  (\bibinfo{year}{2023}).
\newblock


\bibitem[Borgeaud et~al\mbox{.}(2022)]%
        {pmlr-v162-borgeaud22a}
\bibfield{author}{\bibinfo{person}{Sebastian Borgeaud}, \bibinfo{person}{Arthur
  Mensch}, \bibinfo{person}{Jordan Hoffmann}, \bibinfo{person}{Trevor Cai},
  \bibinfo{person}{Eliza Rutherford}, \bibinfo{person}{Katie Millican},
  \bibinfo{person}{George~Bm Van Den~Driessche}, \bibinfo{person}{Jean-Baptiste
  Lespiau}, \bibinfo{person}{Bogdan Damoc}, \bibinfo{person}{Aidan Clark},
  \bibinfo{person}{Diego De~Las~Casas}, \bibinfo{person}{Aurelia Guy},
  \bibinfo{person}{Jacob Menick}, \bibinfo{person}{Roman Ring},
  \bibinfo{person}{Tom Hennigan}, \bibinfo{person}{Saffron Huang},
  \bibinfo{person}{Loren Maggiore}, \bibinfo{person}{Chris Jones},
  \bibinfo{person}{Albin Cassirer}, \bibinfo{person}{Andy Brock},
  \bibinfo{person}{Michela Paganini}, \bibinfo{person}{Geoffrey Irving},
  \bibinfo{person}{Oriol Vinyals}, \bibinfo{person}{Simon Osindero},
  \bibinfo{person}{Karen Simonyan}, \bibinfo{person}{Jack Rae},
  \bibinfo{person}{Erich Elsen}, {and} \bibinfo{person}{Laurent Sifre}.}
  \bibinfo{year}{2022}\natexlab{}.
\newblock \showarticletitle{Improving Language Models by Retrieving from
  Trillions of Tokens}. In \bibinfo{booktitle}{\emph{Proceedings of the 39th
  International Conference on Machine Learning}}
  \emph{(\bibinfo{series}{Proceedings of Machine Learning Research},
  Vol.~\bibinfo{volume}{162})}. \bibinfo{publisher}{PMLR},
  \bibinfo{pages}{2206--2240}.
\newblock


\bibitem[Brown(2020)]%
        {brown2020language}
\bibfield{author}{\bibinfo{person}{Tom~B Brown}.}
  \bibinfo{year}{2020}\natexlab{}.
\newblock \showarticletitle{Language models are few-shot learners}.
\newblock \bibinfo{journal}{\emph{arXiv preprint arXiv:2005.14165}}
  (\bibinfo{year}{2020}).
\newblock


\bibitem[Caruccio et~al\mbox{.}(2024)]%
        {caruccio2024claude}
\bibfield{author}{\bibinfo{person}{Loredana Caruccio}, \bibinfo{person}{Stefano
  Cirillo}, \bibinfo{person}{Giuseppe Polese}, \bibinfo{person}{Giandomenico
  Solimando}, \bibinfo{person}{Shanmugam Sundaramurthy}, {and}
  \bibinfo{person}{Genoveffa Tortora}.} \bibinfo{year}{2024}\natexlab{}.
\newblock \showarticletitle{Claude 2.0 large language model: Tackling a
  real-world classification problem with a new iterative prompt engineering
  approach}.
\newblock \bibinfo{journal}{\emph{Intelligent Systems with Applications}}
  \bibinfo{volume}{21} (\bibinfo{year}{2024}), \bibinfo{pages}{200336}.
\newblock


\bibitem[Chu et~al\mbox{.}(2024)]%
        {chu-etal-2024-navigate}
\bibfield{author}{\bibinfo{person}{Zheng Chu}, \bibinfo{person}{Jingchang
  Chen}, \bibinfo{person}{Qianglong Chen}, \bibinfo{person}{Weijiang Yu},
  \bibinfo{person}{Tao He}, \bibinfo{person}{Haotian Wang},
  \bibinfo{person}{Weihua Peng}, \bibinfo{person}{Ming Liu},
  \bibinfo{person}{Bing Qin}, {and} \bibinfo{person}{Ting Liu}.}
  \bibinfo{year}{2024}\natexlab{}.
\newblock \showarticletitle{Navigate through Enigmatic Labyrinth A Survey of
  Chain of Thought Reasoning: Advances, Frontiers and Future}. In
  \bibinfo{booktitle}{\emph{Proceedings of the 62nd Annual Meeting of the
  Association for Computational Linguistics (Volume 1: Long Papers)}}.
  \bibinfo{publisher}{Association for Computational Linguistics},
  \bibinfo{address}{Bangkok, Thailand}, \bibinfo{pages}{1173--1203}.
\newblock


\bibitem[Dubey et~al\mbox{.}(2024)]%
        {dubey2024llama}
\bibfield{author}{\bibinfo{person}{Abhimanyu Dubey}, \bibinfo{person}{Abhinav
  Jauhri}, \bibinfo{person}{Abhinav Pandey}, \bibinfo{person}{Abhishek Kadian},
  \bibinfo{person}{Ahmad Al-Dahle}, \bibinfo{person}{Aiesha Letman},
  \bibinfo{person}{Akhil Mathur}, \bibinfo{person}{Alan Schelten},
  \bibinfo{person}{Amy Yang}, \bibinfo{person}{Angela Fan}, {et~al\mbox{.}}}
  \bibinfo{year}{2024}\natexlab{}.
\newblock \showarticletitle{The llama 3 herd of models}.
\newblock \bibinfo{journal}{\emph{arXiv preprint arXiv:2407.21783}}
  (\bibinfo{year}{2024}).
\newblock


\bibitem[Edge et~al\mbox{.}(2024)]%
        {edge2024local}
\bibfield{author}{\bibinfo{person}{Darren Edge}, \bibinfo{person}{Ha Trinh},
  \bibinfo{person}{Newman Cheng}, \bibinfo{person}{Joshua Bradley},
  \bibinfo{person}{Alex Chao}, \bibinfo{person}{Apurva Mody},
  \bibinfo{person}{Steven Truitt}, {and} \bibinfo{person}{Jonathan Larson}.}
  \bibinfo{year}{2024}\natexlab{}.
\newblock \showarticletitle{From local to global: A graph rag approach to
  query-focused summarization}.
\newblock \bibinfo{journal}{\emph{arXiv preprint arXiv:2404.16130}}
  (\bibinfo{year}{2024}).
\newblock


\bibitem[Frisoni et~al\mbox{.}(2022)]%
        {frisoni2022bioreader}
\bibfield{author}{\bibinfo{person}{Giacomo Frisoni}, \bibinfo{person}{Miki
  Mizutani}, \bibinfo{person}{Gianluca Moro}, {and} \bibinfo{person}{Lorenzo
  Valgimigli}.} \bibinfo{year}{2022}\natexlab{}.
\newblock \showarticletitle{Bioreader: a retrieval-enhanced text-to-text
  transformer for biomedical literature}. In
  \bibinfo{booktitle}{\emph{Proceedings of the 2022 conference on empirical
  methods in natural language processing}}. \bibinfo{pages}{5770--5793}.
\newblock


\bibitem[He et~al\mbox{.}(2024)]%
        {he2024g}
\bibfield{author}{\bibinfo{person}{Xiaoxin He}, \bibinfo{person}{Yijun Tian},
  \bibinfo{person}{Yifei Sun}, \bibinfo{person}{Nitesh~V Chawla},
  \bibinfo{person}{Thomas Laurent}, \bibinfo{person}{Yann LeCun},
  \bibinfo{person}{Xavier Bresson}, {and} \bibinfo{person}{Bryan Hooi}.}
  \bibinfo{year}{2024}\natexlab{}.
\newblock \showarticletitle{G-retriever: Retrieval-augmented generation for
  textual graph understanding and question answering}.
\newblock \bibinfo{journal}{\emph{arXiv preprint arXiv:2402.07630}}
  (\bibinfo{year}{2024}).
\newblock


\bibitem[Ji et~al\mbox{.}(2022)]%
        {ji2022kg}
\bibfield{author}{\bibinfo{person}{Shaoxiong Ji}, \bibinfo{person}{Shirui Pan},
  \bibinfo{person}{Erik Cambria}, \bibinfo{person}{Pekka Marttinen}, {and}
  \bibinfo{person}{Philip~S. Yu}.} \bibinfo{year}{2022}\natexlab{}.
\newblock \showarticletitle{A Survey on Knowledge Graphs: Representation,
  Acquisition, and Applications}.
\newblock \bibinfo{journal}{\emph{IEEE Transactions on Neural Networks and
  Learning Systems}} \bibinfo{volume}{33}, \bibinfo{number}{2}
  (\bibinfo{year}{2022}), \bibinfo{pages}{494--514}.
\newblock


\bibitem[Ji et~al\mbox{.}(2023)]%
        {ji2023survey}
\bibfield{author}{\bibinfo{person}{Ziwei Ji}, \bibinfo{person}{Nayeon Lee},
  \bibinfo{person}{Rita Frieske}, \bibinfo{person}{Tiezheng Yu},
  \bibinfo{person}{Dan Su}, \bibinfo{person}{Yan Xu}, \bibinfo{person}{Etsuko
  Ishii}, \bibinfo{person}{Ye~Jin Bang}, \bibinfo{person}{Andrea Madotto},
  {and} \bibinfo{person}{Pascale Fung}.} \bibinfo{year}{2023}\natexlab{}.
\newblock \showarticletitle{Survey of Hallucination in Natural Language
  Generation}.
\newblock \bibinfo{journal}{\emph{ACM Comput. Surv.}} \bibinfo{volume}{55},
  \bibinfo{number}{12}, Article \bibinfo{articleno}{248}
  (\bibinfo{year}{2023}), \bibinfo{numpages}{38}~pages.
\newblock


\bibitem[Li et~al\mbox{.}(2023)]%
        {li-etal-2023-shot}
\bibfield{author}{\bibinfo{person}{Tianle Li}, \bibinfo{person}{Xueguang Ma},
  \bibinfo{person}{Alex Zhuang}, \bibinfo{person}{Yu Gu}, \bibinfo{person}{Yu
  Su}, {and} \bibinfo{person}{Wenhu Chen}.} \bibinfo{year}{2023}\natexlab{}.
\newblock \showarticletitle{Few-shot In-context Learning on Knowledge Base
  Question Answering}. In \bibinfo{booktitle}{\emph{Proceedings of the 61st
  Annual Meeting of the Association for Computational Linguistics (Volume 1:
  Long Papers)}}. \bibinfo{pages}{6966--6980}.
\newblock


\bibitem[Li et~al\mbox{.}(2024)]%
        {li2024chainofknowledge}
\bibfield{author}{\bibinfo{person}{Xingxuan Li}, \bibinfo{person}{Ruochen
  Zhao}, \bibinfo{person}{Yew~Ken Chia}, \bibinfo{person}{Bosheng Ding},
  \bibinfo{person}{Shafiq Joty}, \bibinfo{person}{Soujanya Poria}, {and}
  \bibinfo{person}{Lidong Bing}.} \bibinfo{year}{2024}\natexlab{}.
\newblock \showarticletitle{Chain-of-Knowledge: Grounding Large Language Models
  via Dynamic Knowledge Adapting over Heterogeneous Sources}. In
  \bibinfo{booktitle}{\emph{The Twelfth International Conference on Learning
  Representations}}.
\newblock


\bibitem[LUO et~al\mbox{.}(2024)]%
        {luo2024reasoning}
\bibfield{author}{\bibinfo{person}{LINHAO LUO}, \bibinfo{person}{Yuan-Fang Li},
  \bibinfo{person}{Reza Haf}, {and} \bibinfo{person}{Shirui Pan}.}
  \bibinfo{year}{2024}\natexlab{}.
\newblock \showarticletitle{Reasoning on Graphs: Faithful and Interpretable
  Large Language Model Reasoning}. In \bibinfo{booktitle}{\emph{The Twelfth
  International Conference on Learning Representations}}.
\newblock


\bibitem[Shi et~al\mbox{.}(2024)]%
        {shi-etal-2024-replug}
\bibfield{author}{\bibinfo{person}{Weijia Shi}, \bibinfo{person}{Sewon Min},
  \bibinfo{person}{Michihiro Yasunaga}, \bibinfo{person}{Minjoon Seo},
  \bibinfo{person}{Richard James}, \bibinfo{person}{Mike Lewis},
  \bibinfo{person}{Luke Zettlemoyer}, {and} \bibinfo{person}{Wen-tau Yih}.}
  \bibinfo{year}{2024}\natexlab{}.
\newblock \showarticletitle{{REPLUG}: Retrieval-Augmented Black-Box Language
  Models}. In \bibinfo{booktitle}{\emph{Proceedings of the 2024 Conference of
  the North American Chapter of the Association for Computational Linguistics:
  Human Language Technologies (Volume 1: Long Papers)}}.
  \bibinfo{publisher}{Association for Computational Linguistics},
  \bibinfo{address}{Mexico City, Mexico}, \bibinfo{pages}{8371--8384}.
\newblock


\bibitem[Sun et~al\mbox{.}(2024)]%
        {sun2024thinkongraph}
\bibfield{author}{\bibinfo{person}{Jiashuo Sun}, \bibinfo{person}{Chengjin Xu},
  \bibinfo{person}{Lumingyuan Tang}, \bibinfo{person}{Saizhuo Wang},
  \bibinfo{person}{Chen Lin}, \bibinfo{person}{Yeyun Gong},
  \bibinfo{person}{Lionel Ni}, \bibinfo{person}{Heung-Yeung Shum}, {and}
  \bibinfo{person}{Jian Guo}.} \bibinfo{year}{2024}\natexlab{}.
\newblock \showarticletitle{Think-on-Graph: Deep and Responsible Reasoning of
  Large Language Model on Knowledge Graph}. In \bibinfo{booktitle}{\emph{The
  Twelfth International Conference on Learning Representations}}.
\newblock


\bibitem[Talmor and Berant(2018)]%
        {talmor2018web}
\bibfield{author}{\bibinfo{person}{Alon Talmor} {and} \bibinfo{person}{Jonathan
  Berant}.} \bibinfo{year}{2018}\natexlab{}.
\newblock \showarticletitle{The Web as a Knowledge-Base for Answering Complex
  Questions}. In \bibinfo{booktitle}{\emph{Proceedings of the 2018 Conference
  of the North American Chapter of the Association for Computational
  Linguistics: Human Language Technologies, Volume 1 (Long Papers)}}.
  \bibinfo{pages}{641--651}.
\newblock


\bibitem[Team et~al\mbox{.}(2023)]%
        {team2023gemini}
\bibfield{author}{\bibinfo{person}{Gemini Team}, \bibinfo{person}{Rohan Anil},
  \bibinfo{person}{Sebastian Borgeaud}, \bibinfo{person}{Yonghui Wu},
  \bibinfo{person}{Jean-Baptiste Alayrac}, \bibinfo{person}{Jiahui Yu},
  \bibinfo{person}{Radu Soricut}, \bibinfo{person}{Johan Schalkwyk},
  \bibinfo{person}{Andrew~M Dai}, \bibinfo{person}{Anja Hauth},
  {et~al\mbox{.}}} \bibinfo{year}{2023}\natexlab{}.
\newblock \showarticletitle{Gemini: a family of highly capable multimodal
  models}.
\newblock \bibinfo{journal}{\emph{arXiv preprint arXiv:2312.11805}}
  (\bibinfo{year}{2023}).
\newblock


\bibitem[Touvron et~al\mbox{.}(2023)]%
        {touvron2023llama}
\bibfield{author}{\bibinfo{person}{Hugo Touvron}, \bibinfo{person}{Louis
  Martin}, \bibinfo{person}{Kevin Stone}, \bibinfo{person}{Peter Albert},
  \bibinfo{person}{Amjad Almahairi}, \bibinfo{person}{Yasmine Babaei},
  \bibinfo{person}{Nikolay Bashlykov}, \bibinfo{person}{Soumya Batra},
  \bibinfo{person}{Prajjwal Bhargava}, \bibinfo{person}{Shruti Bhosale},
  {et~al\mbox{.}}} \bibinfo{year}{2023}\natexlab{}.
\newblock \showarticletitle{Llama 2: Open foundation and fine-tuned chat
  models}.
\newblock \bibinfo{journal}{\emph{arXiv preprint arXiv:2307.09288}}
  (\bibinfo{year}{2023}).
\newblock


\bibitem[Wang et~al\mbox{.}(2017)]%
        {wang2017kg}
\bibfield{author}{\bibinfo{person}{Quan Wang}, \bibinfo{person}{Zhendong Mao},
  \bibinfo{person}{Bin Wang}, {and} \bibinfo{person}{Li Guo}.}
  \bibinfo{year}{2017}\natexlab{}.
\newblock \showarticletitle{Knowledge Graph Embedding: A Survey of Approaches
  and Applications}.
\newblock \bibinfo{journal}{\emph{IEEE Transactions on Knowledge and Data
  Engineering}} \bibinfo{volume}{29}, \bibinfo{number}{12}
  (\bibinfo{year}{2017}), \bibinfo{pages}{2724--2743}.
\newblock


\bibitem[Wang et~al\mbox{.}(2023)]%
        {wang2023retrievalbased}
\bibfield{author}{\bibinfo{person}{Zichao Wang}, \bibinfo{person}{Weili Nie},
  \bibinfo{person}{Zhuoran Qiao}, \bibinfo{person}{Chaowei Xiao},
  \bibinfo{person}{Richard Baraniuk}, {and} \bibinfo{person}{Anima
  Anandkumar}.} \bibinfo{year}{2023}\natexlab{}.
\newblock \showarticletitle{Retrieval-based Controllable Molecule Generation}.
  In \bibinfo{booktitle}{\emph{The Eleventh International Conference on
  Learning Representations}}.
\newblock


\bibitem[Wei et~al\mbox{.}(2022a)]%
        {wei2022finetuned}
\bibfield{author}{\bibinfo{person}{Jason Wei}, \bibinfo{person}{Maarten Bosma},
  \bibinfo{person}{Vincent Zhao}, \bibinfo{person}{Kelvin Guu},
  \bibinfo{person}{Adams~Wei Yu}, \bibinfo{person}{Brian Lester},
  \bibinfo{person}{Nan Du}, \bibinfo{person}{Andrew~M. Dai}, {and}
  \bibinfo{person}{Quoc~V Le}.} \bibinfo{year}{2022}\natexlab{a}.
\newblock \showarticletitle{Finetuned Language Models are Zero-Shot Learners}.
  In \bibinfo{booktitle}{\emph{International Conference on Learning
  Representations}}.
\newblock


\bibitem[Wei et~al\mbox{.}(2022b)]%
        {wei2022cot}
\bibfield{author}{\bibinfo{person}{Jason Wei}, \bibinfo{person}{Xuezhi Wang},
  \bibinfo{person}{Dale Schuurmans}, \bibinfo{person}{Maarten Bosma},
  \bibinfo{person}{brian ichter}, \bibinfo{person}{Fei Xia},
  \bibinfo{person}{Ed Chi}, \bibinfo{person}{Quoc~V Le}, {and}
  \bibinfo{person}{Denny Zhou}.} \bibinfo{year}{2022}\natexlab{b}.
\newblock \showarticletitle{Chain-of-Thought Prompting Elicits Reasoning in
  Large Language Models}. In \bibinfo{booktitle}{\emph{Advances in Neural
  Information Processing Systems}}, Vol.~\bibinfo{volume}{35}.
  \bibinfo{pages}{24824--24837}.
\newblock


\bibitem[Wu et~al\mbox{.}(2024)]%
        {wu2024stark}
\bibfield{author}{\bibinfo{person}{Shirley Wu}, \bibinfo{person}{Shiyu Zhao},
  \bibinfo{person}{Michihiro Yasunaga}, \bibinfo{person}{Kexin Huang},
  \bibinfo{person}{Kaidi Cao}, \bibinfo{person}{Qian Huang},
  \bibinfo{person}{Vassilis~N Ioannidis}, \bibinfo{person}{Karthik Subbian},
  \bibinfo{person}{James Zou}, {and} \bibinfo{person}{Jure Leskovec}.}
  \bibinfo{year}{2024}\natexlab{}.
\newblock \showarticletitle{STaRK: Benchmarking LLM Retrieval on Textual and
  Relational Knowledge Bases}.
\newblock \bibinfo{journal}{\emph{arXiv preprint arXiv:2404.13207}}
  (\bibinfo{year}{2024}).
\newblock


\bibitem[Xu et~al\mbox{.}(2024)]%
        {xu2024generate}
\bibfield{author}{\bibinfo{person}{Yao Xu}, \bibinfo{person}{Shizhu He},
  \bibinfo{person}{Jiabei Chen}, \bibinfo{person}{Zihao Wang},
  \bibinfo{person}{Yangqiu Song}, \bibinfo{person}{Hanghang Tong},
  \bibinfo{person}{Kang Liu}, {and} \bibinfo{person}{Jun Zhao}.}
  \bibinfo{year}{2024}\natexlab{}.
\newblock \showarticletitle{Generate-on-Graph: Treat LLM as both Agent and KG
  in Incomplete Knowledge Graph Question Answering}.
\newblock \bibinfo{journal}{\emph{arXiv preprint arXiv:2404.14741}}
  (\bibinfo{year}{2024}).
\newblock


\bibitem[Yang et~al\mbox{.}(2024)]%
        {yang2024qwen2}
\bibfield{author}{\bibinfo{person}{An Yang}, \bibinfo{person}{Baosong Yang},
  \bibinfo{person}{Binyuan Hui}, \bibinfo{person}{Bo Zheng},
  \bibinfo{person}{Bowen Yu}, \bibinfo{person}{Chang Zhou},
  \bibinfo{person}{Chengpeng Li}, \bibinfo{person}{Chengyuan Li},
  \bibinfo{person}{Dayiheng Liu}, \bibinfo{person}{Fei Huang}, {et~al\mbox{.}}}
  \bibinfo{year}{2024}\natexlab{}.
\newblock \showarticletitle{Qwen2 technical report}.
\newblock \bibinfo{journal}{\emph{arXiv preprint arXiv:2407.10671}}
  (\bibinfo{year}{2024}).
\newblock


\bibitem[Yao et~al\mbox{.}(2023)]%
        {yao2023react}
\bibfield{author}{\bibinfo{person}{Shunyu Yao}, \bibinfo{person}{Jeffrey Zhao},
  \bibinfo{person}{Dian Yu}, \bibinfo{person}{Nan Du}, \bibinfo{person}{Izhak
  Shafran}, \bibinfo{person}{Karthik~R Narasimhan}, {and} \bibinfo{person}{Yuan
  Cao}.} \bibinfo{year}{2023}\natexlab{}.
\newblock \showarticletitle{ReAct: Synergizing Reasoning and Acting in Language
  Models}. In \bibinfo{booktitle}{\emph{The Eleventh International Conference
  on Learning Representations}}.
\newblock


\bibitem[Yih et~al\mbox{.}(2016)]%
        {yih2016value}
\bibfield{author}{\bibinfo{person}{Wen-tau Yih}, \bibinfo{person}{Matthew
  Richardson}, \bibinfo{person}{Christopher Meek}, \bibinfo{person}{Ming-Wei
  Chang}, {and} \bibinfo{person}{Jina Suh}.} \bibinfo{year}{2016}\natexlab{}.
\newblock \showarticletitle{The value of semantic parse labeling for knowledge
  base question answering}. In \bibinfo{booktitle}{\emph{Proceedings of the
  54th Annual Meeting of the Association for Computational Linguistics (Volume
  2: Short Papers)}}. \bibinfo{pages}{201--206}.
\newblock


\bibitem[Zhang et~al\mbox{.}(2023a)]%
        {zhang-etal-2023-repocoder}
\bibfield{author}{\bibinfo{person}{Fengji Zhang}, \bibinfo{person}{Bei Chen},
  \bibinfo{person}{Yue Zhang}, \bibinfo{person}{Jacky Keung},
  \bibinfo{person}{Jin Liu}, \bibinfo{person}{Daoguang Zan},
  \bibinfo{person}{Yi Mao}, \bibinfo{person}{Jian-Guang Lou}, {and}
  \bibinfo{person}{Weizhu Chen}.} \bibinfo{year}{2023}\natexlab{a}.
\newblock \showarticletitle{{R}epo{C}oder: Repository-Level Code Completion
  Through Iterative Retrieval and Generation}. In
  \bibinfo{booktitle}{\emph{Proceedings of the 2023 Conference on Empirical
  Methods in Natural Language Processing}}. \bibinfo{publisher}{Association for
  Computational Linguistics}, \bibinfo{address}{Singapore},
  \bibinfo{pages}{2471--2484}.
\newblock


\bibitem[Zhang et~al\mbox{.}(2023b)]%
        {zhang2023automatic}
\bibfield{author}{\bibinfo{person}{Zhuosheng Zhang}, \bibinfo{person}{Aston
  Zhang}, \bibinfo{person}{Mu Li}, {and} \bibinfo{person}{Alex Smola}.}
  \bibinfo{year}{2023}\natexlab{b}.
\newblock \showarticletitle{Automatic Chain of Thought Prompting in Large
  Language Models}. In \bibinfo{booktitle}{\emph{The Eleventh International
  Conference on Learning Representations}}.
\newblock


\bibitem[Zhao et~al\mbox{.}(2024)]%
        {Zhao2024RetrievalAugmentedGF}
\bibfield{author}{\bibinfo{person}{Penghao Zhao}, \bibinfo{person}{Hailin
  Zhang}, \bibinfo{person}{Qinhan Yu}, \bibinfo{person}{Zhengren Wang},
  \bibinfo{person}{Yunteng Geng}, \bibinfo{person}{Fangcheng Fu},
  \bibinfo{person}{Ling Yang}, \bibinfo{person}{Wentao Zhang}, {and}
  \bibinfo{person}{Bin Cui}.} \bibinfo{year}{2024}\natexlab{}.
\newblock \showarticletitle{Retrieval-Augmented Generation for AI-Generated
  Content: A Survey}.
\newblock \bibinfo{journal}{\emph{ArXiv}}  \bibinfo{volume}{abs/2402.19473}
  (\bibinfo{year}{2024}).
\newblock


\end{thebibliography}

% \newpage
\appendix

\section{Prompts}
\label{app:prompt}

Solve a question answering task with interleaving Thought, Action, Observation steps. Thought can reason about the current situation, and Action can be three types:

(1) Search[node1 | node2 | ...], which searches the exact nodes on the knowledge graph and returns their one-hop subgraphs. You should extract the all concrete nodes appeared in your last thought without redundant words, and you should always select nodes from topic nodes in the first search.

(2) Query[question], which finds the most related node of the given question based on text embedding similarity.

(3) Finish[answer1 | answer2 | ...], which returns the answer and finishes the task. The answers should be complete node name appeared in the triples. If you don't know the answer, please output Finish[unknown].

Nodes and answers should be separated by tab.

You should generate each step without redundant words.

Here are some examples

$\cdots$

Question: ...

Topic Node: [...]

Thought 1: ...

Action 1: Search/Query[...]

Observation 1: ...

Thought 2: ...

Action 2: Search/Query[...]

Observation 2: ...

...

Thought $n$: ...

Action $n$: Finish[...]

\section{Experiment Details}
\label{app:data}

In our experiments, 
we consider the following three semi-structured data sets from~\cite{wu2024stark}: 
Amazon, MAG and PrimeKG. 
Amazon is a product recommendation data set. 
% The dataset emphasizes customer-centric factors like product quality, functionality, and style 
and incorporates single-hop relational data involving brands, categories, and products in complementary or substitute relationships.
% Queries are presented in a conversational style, making the data more relevant to practical scenarios. 
MAG is an academic data set featuring a complex network of nodes and edges centered around paper nodes, particularly focusing on citation and authorship. 
The queries involve single-hop or multi-hop relational queries along with textual properties sourced primarily from abstracts, such as the paper's topic and methodology. 
PrimeKG is a biomedical data set related to diseases, drugs, genes/proteins, etc. 
% It features multi-hop queries that incorporate diverse biomedical properties and ensure each query reflects a real-world scenario. Furthermore, PrimeKG enhances query diversity by adopting three perspectives — medical scientist, doctor, and patient — during query generation to comprehensively evaluate retrieval systems. 
Detailed statistics of these semi-structured data is in Table~\ref{tab:statistics}

\begin{table}[ht]
    \centering
    \caption{Data statistics of semi-structured data used in experiments}
    \resizebox{1.0\columnwidth}{!}{
    \begin{tabular}{l|ccrrrrrrr}
        \toprule
        \multirow{ 2}{*}{} & {\#node} & {\#edge} & {avg. } & \multirow{ 2}{*}{{\#nodes}} & \multirow{ 2}{*}{{\#edges}} & \multirow{ 2}{*}{{\#queries}} & \multirow{ 2}{*}{{train/val/test}} \\
        & {types} & {types} & {degree} \\
        \midrule
        Amazon & 4 & 4  & 18.2 & 1,035,542 & 9,443,802 & 9,100 & 0.65/0.17/0.18 \\
        MAG & 4 & 4  & 43.5& 1,872,968 & 39,802,116 & 13.323 & 0.60/0.20/0.20 \\
        PrimeKG & 10& 18 & 125.2 & 129,375 & 8,100,498 & 11,204 & 0.55/0.20/0.25 \\
        \bottomrule
    \end{tabular}
    }
    % \hspace{3pt}
    \label{tab:statistics}
\end{table}

\section{More Experiment Results}
\label{app:exp}

Table~\ref{tab:f1} compares the macro F1 score of different methods. 
Note that F1 score is not applicable to RAG-based methods (VSS and VSS+GPT-4 reranker) as these two methods only output a ranked list for all nodes, and does not specify which nodes will be the final answers. 
As such, we can only compute the F1 score for KGQA-based methods. 
Here we use the macro F1 score as the micro F1 score should be exactly the same as the Hit@1 score, and we can see that the proposed method still outperforms other baseline when using the macro F1 score as performance metric. 

\begin{table}[t]
  \caption{F1 scores on different semi-structured data with different methods. Best highlighted in bold.}
  \label{tab:f1}
  \begin{tabular}{c c c c c}
    \toprule
    % Method & MRR & Hits@1 & Hits@3 & Hits@10 \\
    % \midrule
    % RotatE & 0.279 & 0.184 & 0.248 & 0.432 \\
    % NBFNet & 0.177 & 0.112 & 0.168 & 0.296 \\
    dataset	& PrimeKG & MAG & Amazon \\
    \midrule
    VSS & 0.1263 & 0.2908 & 0.4044 \\
    VSS+GPT-4 reranker & 0.1828 & 0.4090 & 0.4479 \\
    \midrule
    ToG & 0.1265 & 0.0528 & 0.1639 \\
    RoG & 0.2216 & 0.3267 & 0.3097 \\
    GoG & 0.1832 & 0.4105 & 0.2079 \\
    \midrule
    PASemiQA (GPT-4 Agent) & {\bf 0.2898} & {\bf 0.4294} & {\bf 0.4542} \\
    \bottomrule
    \end{tabular}
\end{table}

Since KGQA methods only output several nodes that are likely to be the answer instead of a complete ranking on all nodes, 
computing Hit@k or Recall@k for arbitrary $k$ is often not possible and may cause unfair comparison. 
Table~\ref{tab:mrr} compares the MRR scores for different baseline methods, and our method PASemiQA generally achieves the best performance under this metric. 

\begin{table}[t]
  \caption{MRR scores on different semi-structured data with different methods. Best highlighted in bold.}
  \label{tab:mrr}
  \begin{tabular}{c c c c c}
    \toprule
    % Method & MRR & Hits@1 & Hits@3 & Hits@10 \\
    % \midrule
    % RotatE & 0.279 & 0.184 & 0.248 & 0.432 \\
    % NBFNet & 0.177 & 0.112 & 0.168 & 0.296 \\
    dataset	& PrimeKG & MAG & Amazon \\
    \midrule
    VSS & 0.2141 & 0.3862 & 0.5035 \\
    VSS+GPT-4 reranker & 0.2655 & 0.4932 & 0.5534 \\
    \midrule
    ToG & 0.1584 & 0.0614 & 0.2118 \\
    RoG & 0.2516 & 0.3867 & 0.3596 \\
    GoG & 0.2108 & 0.4783 & 0.2628 \\
    \midrule
    PASemiQA (GPT-4 Agent) & {\bf 0.3102} & {\bf 0.5024} & {\bf 0.5568} \\
    \bottomrule
    \end{tabular}
\end{table}

\end{document}